\pdfoutput=1

\documentclass[11pt]{article}

\usepackage[]{acl}
\usepackage{booktabs}
\usepackage{CJKutf8}
\usepackage{times}
\usepackage{amsmath}
\usepackage{amssymb}
\usepackage{amsthm}
\usepackage{latexsym}
\usepackage{graphicx} 
\usepackage{textcomp}
\usepackage{enumitem}
\usepackage[T1]{fontenc}
\usepackage{tcolorbox}
\usepackage{wrapfig}
\usepackage{setspace}
\usepackage{parskip}

\usepackage[utf8]{inputenc}

\usepackage{microtype}

\usepackage{inconsolata}

\title{Pride and Prejudice: LLM Amplifies Self-Bias in Self-Refinement}

\author{
Wenda Xu$^\dagger$, Guanglei Zhu$^\ddag$, Xuandong Zhao$^\dag$, Liangming Pan$^\dag$, \\
\textbf{Lei Li$^\ddag$, William Yang Wang$^\dag$}\\
$^\dagger$University of California, Santa Barbara,
$^\ddag$Carnegie Mellon University\\
\texttt{\{wendaxu,xuandongzhao,liangmingpan,william\}@cs.ucsb.edu},\\ \texttt{\{guanglez,leili\}@cs.cmu.edu}
}
\begin{document}
\maketitle
\begin{abstract}

Recent studies show that large language models (LLMs) improve their performance through self-feedback on certain tasks while degrade on others. We discovered that such a contrary is due to LLM's bias in evaluating their own output.  
In this paper, we formally define LLM's self-bias -- the tendency to favor its own generation -- using two statistics. We analyze six LLMs (GPT-4, GPT-3.5, Gemini, LLaMA2, Mixtral and DeepSeek) on translation, constrained text generation, and mathematical reasoning tasks. We find that self-bias is prevalent in all examined LLMs across multiple languages and tasks. Our analysis reveals that while the self-refine pipeline improves the fluency and understandability of model outputs, it further amplifies self-bias. To mitigate such biases, we discover that larger model size and external feedback with accurate assessment can significantly reduce bias in the self-refine pipeline, leading to actual performance improvement in downstream tasks. The code and data are released at \url{https://github.com/xu1998hz/llm\_self\_bias}.

\end{abstract}

\section{Introduction}

\begin{figure}[t]
    \centering
    \includegraphics[width=\linewidth]{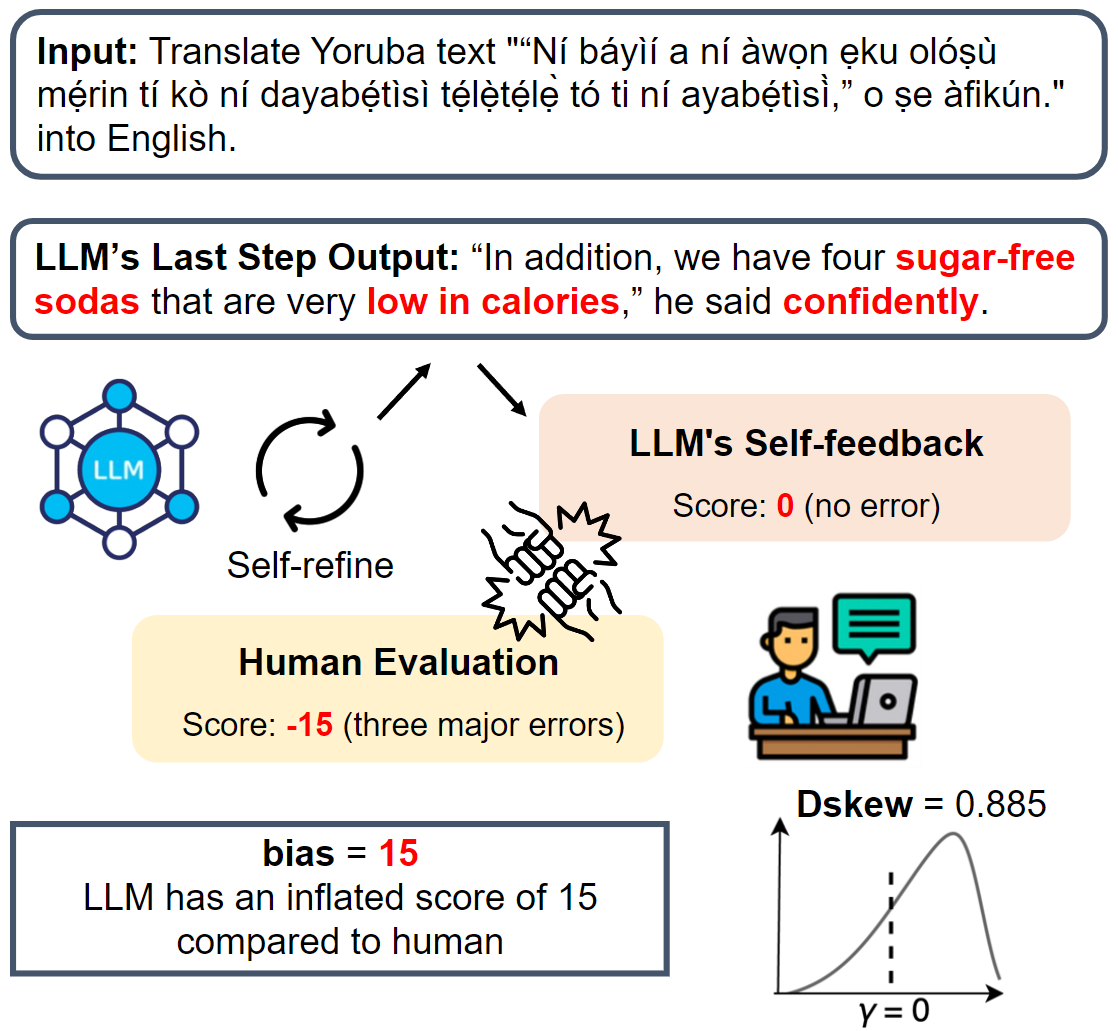}
    \caption{How LLM's self-feedback inflates scores compared to human assessment. Bias is the mean difference between LLM and human scores, while skewness (Dskew) measures the asymmetry of their distribution around zero. Non-biased estimation will have Dskew=0.}
    \label{fig:pipeline}
\end{figure}

Large language models (LLMs) have shown strong capabilities in many NLP tasks. 
While these models still make mistakes, recent studies show that ``self-refine'' (also known as ``self-reflection'') is promising to rectify errors based on LLM's self-feedback~\cite{madaan2023selfrefine,chen2023teaching,shinn2023reflexion,manakul2023selfcheckgpt,DBLP:Self-Correction-Survey}. 
Meanwhile, opposite study also shows that LLMs fail to correct their mistakes and their performance even gets worse after self-feedback~\cite{DBLP:llm_not_self_correct}. 
These contradictory results suggest that LLM's self-feedback is unreliable. 
Self-refine procedure relies on LLM's evaluation capability of the generated text. 
We hypothesize that if there is a bias during the self-evaluation process, such bias will be amplified during iterative self-refinement. 
This is consistent with a prior finding that LM-based metrics (e.g. BARTScore) exhibit ``narcissism'' during self-evaluation, \textit{i.e.}, the metric model favors text generated by the same underlying language model in the context of summarization tasks~\cite{liu2023llms}. 
However, it remains unclear whether bias exists universally in LLMs across a wide range of tasks. 
How to quantify such biases? How does this ``narcissism'' impact LLM's self-refinement?





In this work, we define ``self-bias'' to the degree that an LLM favors its own generation. 
We propose to use two principled statistics to estimate self-bias in LLM's self-refinement procedure. 
The first one measures the degree of inflation in the LLM's self-evaluation compared to the true (human) evaluation.  
The second measures whether LLM's self-evaluation is skewed compared to the ture estimate. 
Figure~\ref{fig:pipeline} illustrates these two statistics. 
We examine self-bias scores on six diverse LLMs, covering four languages across three distinct tasks: machine translation, constrained text generation, and mathematical reasoning. 
We find that self-bias is universal in self-refine and self-rewarding pipelines, regardless of the languages and tasks. 
This bias causes LLMs to optimize for false positive corrections rather than improving the actual output quality. 

We further investigate what is the real benefit of self-refine. 
We find that while the self-refine pipeline improves the fluency and understandability of model outputs, it does not necessarily lead to intended improvements as specified in the prompt. Moreover, LLMs may favor texts that mirror their style, potentially leading to false positive optimization and reduced diversity in text generation. 
To mitigate the self-bias, we propose two solutions: increasing the model size and incorporating external feedback to provide accurate assessment, thereby directing the LLM towards more accurate self-correction. Our contributions are: 
\begin{enumerate}[leftmargin=*, itemsep=0pt, topsep=2pt]
    \item We formally define the self-bias of an LLM using two principled estimated statistics.
    \item We quantify self-biases for six diverse LLMs and find that self-bias amplifies during self-refine across many languages and tasks. 
    \item We observe two factors that contribute to self-bias and pinpoint two directions to mitigate it and elicit LLMs' self-correction ability. 
\end{enumerate}

\section{Related Work}


\paragraph{Large Language Model Self-correction.}
Recent works demonstrate that LLM can utilize its own feedback signal to refine itself~\cite{madaan2023selfrefine, chen2023teaching, shinn2023reflexion}. 
\citet{wang2023selfconsistency} further proposed to sample diverse reasoning paths and use a majority vote to find the most confident answer. \citet{huang-etal-2023-large} leverages self-consistency to further fine-tune the LLM on the most confident reasoning path with diverse instruction formats. On the other hand, LLM's self-feedback can also be used as a reward signal to further align LLM to follow instructions \cite{gulcehre2023reinforced, yuan2024selfrewarding}. 

Despite some demonstrations of performance improvements, most findings indicate that LLMs struggle to rectify their initial mistakes, and their performance even worsens after self-correction~\cite{DBLP:llm_not_self_correct,DBLP:llm_cannot_find, DBLP:CritiqueLLM}. This issue arises because the quality of the model's self-generated feedback is bounded by its existing knowledge and abilities \cite{DBLP:gpt4_wrong, DBLP:a_closer_look}. Therefore, internal feedback may not offer any extra advantage for improving the results; it might even steer the model away from the correct answer \cite{DBLP:can_llm_improve}. However, prior works only had empirical observations on this phenomenon, while lacking a quantitative analysis. Moreover, prior works only focus on specific tasks, such as reasoning or code generation. In this work, we are the first to quantitatively analyze the self-bias of different LLMs across three tasks and four languages, which provides a novel and generalizable view to address the perils of self-refine.  


\paragraph{LLMs as Evaluators.}
\citet{liu2023geval} leverages GPT-4 to evaluate text through chain-of-thoughts prompting. \citet{fu2023gptscore} leverages GPT-3's sequence likelihood to estimate model performance. \citet{kocmi-federmann-2023-gemba, xu-etal-2023-instructscore} designed detailed error schemes for LLM to output fine-grained error annotations. 
Despite the popularity of using LLMs as evaluators, \citet{koo2023benchmarking} pointed out that LLM exhibits cognitive bias when evaluating the text, misaligning from human preference. \citet{zheng2023judging} pointed out LLMs have verbosity and self-enhancement bias, which makes them prefer long and verbose answers and answers generated by themselves. \citet{chang-etal-2023-speak} found out that LLM prefers memorized text over non-memorized text, creating unfair judgments over texts. \citet{deutsch-etal-2022-limitations, liu2023llms} point out that reference-free metrics are inherently biased on their own outputs. 

Although the above empirical studies provide valuable insights, they lack a formal definition to quantify those biases nor provide a connection to the self-refine framework. In this work, we define and quantify self-bias and provide the first in-depth analysis of its impact on the self-refine pipeline. We analyze potential bias attributions and pinpoint two mitigation directions.


\section{Quantifying Self-Bias}

This section outlines the approach used to quantify the self-bias exhibited by LLMs in an iterative self-refinement pipeline. We employ statistical bias and distance skewness \cite{asymetry} estimation to measure self-bias. 

\subsection{Iterative Self-Refinement in LLMs}

Self-refinement is an inference time method, in which the LLM first generates a response $y_i$ to a given prompt $x$, and then the same LLM generates feedback $f_i$ based on the candidate output $y_i$ and input $x$. Based on feedback $f_i$, input $x$, and candidate output $y_i$, the LLM then generates a refined output $r_i$. LLM iterates between the feedback and the refinement steps, continuing until it reaches a predetermined number of iterations. 
At each refinement step, the refined output will only be accepted if it demonstrates superior quality compared to the previously generated text. The quality of the text is assessed through self-feedback from the language model itself. At each feedback or refinement step, LLM only sees the last iteration’s generation or feedback, without accessing the entire history of output or feedback. 

\subsection{Bias Estimation}
We estimate the self-bias of LLMs using the statistical bias definition. This bias is characterized by the disparity between an LLM's predicted quality score and the expected quality score, as follows:
\begin{equation}
\textrm{Bias}(\hat{\theta}) = \frac{1}{n}\sum_{i=1}^{n} (\mathbb{E}[\hat{\theta}_i]-\theta_i) ,
\end{equation}
where $\mathbb{E}[\hat{\theta}_i]$ is an expected LLM's quality prediction at sample $i$, and $\theta_i$ denotes the true quality of sample $i$.
Ideally, $\theta_i$ should be derived from human annotations, for example, multidimensional quality metrics (MQM) human annotations \cite{freitag-etal-2021-experts} for machine translation, or predefined criteria such as word coverage for constrained text generation \cite{madaan2023selfrefine}. The LLM's quality prediction is expected to precisely follow the human annotation procedure or predefined criteria, ensuring consistency between $\theta$ and $\mathbb{E}[\hat{\theta}]$. When $\textrm{Bias}(\hat{\theta}) > 0$, the LLM assigns a higher quality score to its own sample compared to the expected quality score. When $\textrm{Bias}(\hat{\theta}) < 0$, the LLM underestimates the sample quality compared to the expected quality score. The larger the value of $\textrm{Bias}(\hat{\theta})$, the more pronounced the LLM's bias against its own samples. 

\begin{figure}[t]
  \centering
    \includegraphics[width=0.9\linewidth]{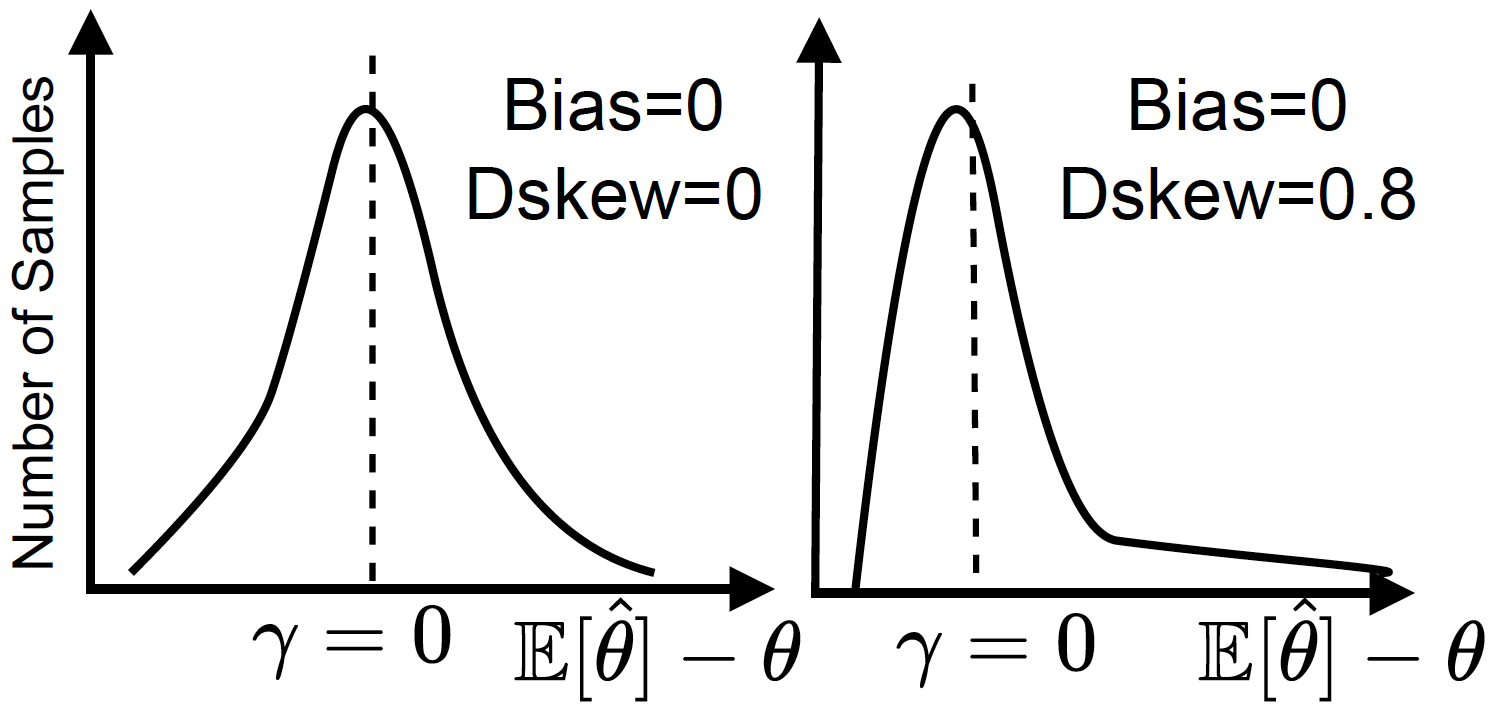}
  \caption{$\textrm{Bias}(\hat{\theta})=0$ does not guarantee a symmetric distribution of $\mathbb{E}[\hat{\theta}]-\theta$. One tail could be long and thin, while the other is short and fat (shown in the right figure). We use distance skewness to measure the asymmetry of distribution. Therefore, using two meta-metrics as complimentary, we can measure the self-bias of LLM.}
\label{fig:compare}
\end{figure}

\subsection{Distance Skewness Estimation}
In an ideal scenario, an unbiased LLM should have equal chance of over-estimation and under-estimation of text quality ($\textrm{Bias}(\hat{\theta})=0$), resulting in a perfectly symmetric distribution when plotting $\mathbb{E}[\hat{\theta}]-\theta$. However, $\textrm{Bias}(\hat{\theta})=0$ does not guarantee a symmetric distribution (In Figure \ref{fig:compare}, one tail could be long and thin, while the other is short and fat, yet they balance out overall). Therefore, we introduce another meta-metric, distance skewness, to measure the asymmetry of $\mathbb{E}[\hat{\theta}]-\theta$'s distribution. Specifically, 
\begin{equation}
d\textrm{Skew}_n(X) = 1 - \frac{\sum_{i,j} \|x_i-x_j\| }{\sum_{i,j} \|x_i+x_j-2\gamma \|},
\end{equation}
where $x_i$ and $x_j$ are two independent identical random examples drawn from $\mathbb{E}[\hat{\theta}]-\theta$. $d\textrm{Skew}_n(X)$ measures the asymmetry of $X$ with respect to $\gamma$. Distance skewness ranges between $0$ and $1$. $d\textrm{Skew}_n(X)$ equals $0$ if and only if $X$ is diagonally distributed respect to $\gamma$. 
$d\textrm{Skew}_n(X)$ equals $1$ if and only if $X$ is distributed at a constant on one side of $\gamma$. A higher distance skewness indicates a more asymmetric distribution of $\mathbb{E}[\hat{\theta}]-\theta$. In our experimental setup, we use both bias and distance skewness to measure the model's bias towards its quality prediction. 

\section{Analyzing LLM's Self-Bias}

\subsection{Experimental Setup}
\label{sec:setup_details}
We include three closed-source LLMs (GPT-4 \cite{openai2023gpt4}, GPT-3.5-Turbo and Gemini \cite{geminiteam2023gemini}) and three open-source LLMs (LLaMA2-7B~\cite{touvron2023llama}, Mixtral-MOE 8x7B \cite{jiang2024mixtral} and DeepSeekMoE 16B \cite{dai2024deepseekmoe}). These models have been shown to have strong instruction-following capabilities \cite{madaan2023selfrefine, shinn2023reflexion}, making them well-suited to demonstrate self-bias. 

For each model, we first prompt it to produce the initial generation. Then, we prompt the model to generate the feedback for the initial generation. The model takes in both the feedback and the prior step generation to produce a refined output. 
We will only accept refinement if the feedback score is improved on the refined output. We listed specific model API/checkpoints in Appendix Section \ref{sec:model_ckpt}.

\paragraph{Machine Translation.} We evaluated LLMs on Flores-200 \cite{nllb2022} dataset with four language pairs: Yoruba to English (Yor-En), Javanese to English (Jav-En), Armenian to English (Arm-En), and Igbo to English (Ig-En), using 100 test examples per language pair. We concentrate on low-to-medium resource language pairs, as \citet{kocmi-etal-2023-findings} indicate that LLMs like GPT-4 already perform at a nearly human-like level in high resource language pairs such as Chinese-to-English, leaving limited potential for further improvement through self-refine. 

To ensure high-quality evaluations, we utilized feedback prompts based on the MQM human annotation from \citet{freitag-etal-2021-experts}, as in \citet{kocmi-federmann-2023-gemba}. LLMs will input source text and candidate text and output feedback, including error location, error type, and severity labels. We adopt the same error scoring as \citet{freitag-etal-2021-experts}, assigning $-1$ for minor errors and $-5$ for major errors, with a score range of 0 to $-25$ (0 for perfect translations, $-25$ for samples with more than five severe errors). The details of the prompts are provided in the Appendix Table \ref{tab:mt_init}, \ref{tab:mt_feedback} and \ref{tab:mt_refinement}. 

Ideally, human raters would have evaluated each sample, but due to cost and scalability constraints, we utilized the reference-based learned metric BLEURT
\cite{sellam2020bleurt} as an approximation of human judgments. BLEURT generates quality scores based on the similarity between candidate and reference translations. To align BLEURT's score distribution with that of human ratings, we employed quantile mapping \cite{BiasCorrectionofGCMPrecipitationbyQuantileMappingHowWellDoMethodsPreserveChangesinQuantilesandExtremes}, yielding a score range from 0 to -25. Although automatic metrics are primarily used, we also conduct modified MQM human evaluations \cite{freitag-etal-2021-experts} for validation purposes. Our bias estimation ranged from -25 to 25. Details on quantile mapping are provided in the Appendix Section \ref{sec:quantile_mapping}. 


\paragraph{Constrained Text Generation.} We conducted experiments on commonsense text generation, following \cite{lin-etal-2020-commongen}. We tested LLMs on 100 examples from the CommonGen Hard dataset. For each testing instance, the large language model (LLM) received approximately 30 concepts and was tasked with generating a fluent and logically sound text. To generate the initial output, we adopted a similar prompt design to that of \cite{lin-etal-2020-commongen}. Next, we provided two ICL feedback examples to help the LLM identify missing concepts in its initial output. In each feedback example, the LLM was given concept words and the previous generation and asked to indicate any missing concepts. This feedback allowed the LLM to revise its output and generate a text with better coverage of the input concepts. The details of the prompts are included in the Appendix Table \ref{tab:commongen_init}, \ref{tab:commongen_feedback} and \ref{tab:commongen_refine}.

To evaluate the coverage of the generated texts, we adopted the evaluation metric used in \cite{madaan2023selfrefine}. This metric uses strict string matching to determine whether each concept word from the input appears in the generated text (metric outputs $1$ if all concepts are covered and $0$ otherwise). From feedback of LLM's missing concepts, we assigned a binary score (0 or 1) to each text based on its full coverage of concepts. Since our string-matching metric and LLM feedback score were on the same scale, we were able to compute bias and distance skewness directly. The range of bias estimation is between $-1$ to $1$. 


\paragraph{Mathematical Reasoning.} We conducted experiments on mathematical reasoning. We tested LLMs on 100 examples from the MATH testing set \cite{hendrycksmath2021}. For each instance, LLM receives a problem statement and generates a step-by-step solution with a final answer. In this task, we use the self-refine pipeline by providing the feedback on the step-by-step solution. In each iteration, the previous solution will be compared against the ground truth answer, outputting $1$  if they are matched and $0$ otherwise. Therefore, we can directly compute bias and distance skewness. The range of bias estimation is between $-1$ to $1$. The details of the prompts are included in the Appendix Table \ref{tab:math_init}. In addition, we also conducted experiments by replacing the self-evaluation (LLM as evaluator) with self-consistency verification (self-consistency as an evaluator) \cite{huang-etal-2023-large}. We include those results in the Appendix \ref{sec:self_consistency_math}. 

\subsection{Self-Bias Amplification during Iterative Refinement}
\begin{figure}[t]
  \centering
    \includegraphics[width=\linewidth]{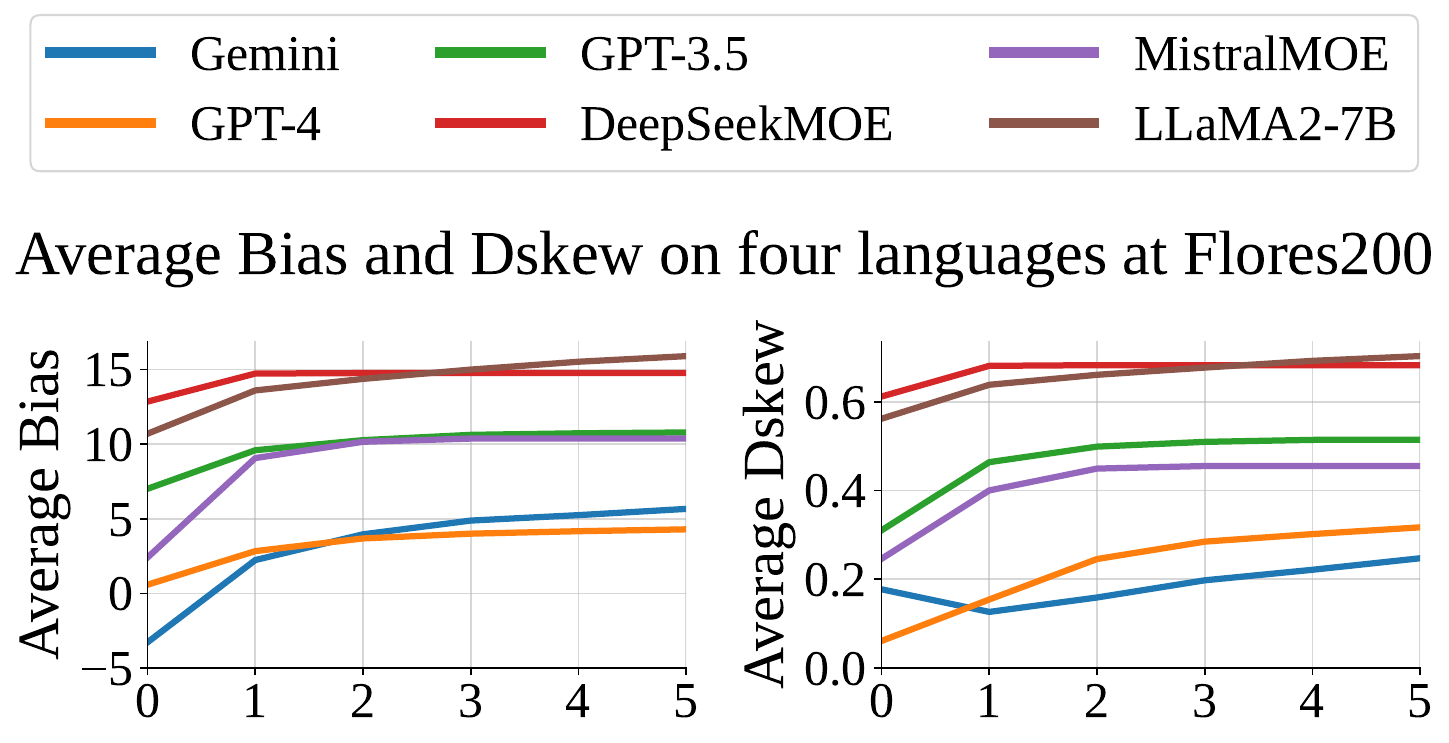}
  
  \caption{Average Bias and Dskew estimations for Yor-En, Jav-En, Arm-En, and Ig-En translations on FLores200, with the $x$-axis showing self-refine steps, reveal that all LLMs exhibit self-bias, where open-source LLMs exhibit higher levels than GPT-4 and Gemini.}
  \label{fig:bias_skewness_flores200}
\end{figure}

\paragraph{Machine Translation.}
In Figure \ref{fig:bias_skewness_flores200}, we illustrate that all large language models (LLMs) exhibit a self-bias in the self-refine pipeline. Notably, open-source LLMs and GPT-3.5-Turbo tend to exhibit higher levels of self-bias throughout iterations than stronger instruction-following LLMs, such as GPT-4 and Gemini. This suggests that GPT-4 and Gemini possess a certain level of capability in resisting self-bias. However, despite some robustness demonstrated by GPT-4 and Gemini, we observe a consistent amplification of self-bias through the self-refine pipeline across four language directions, indicating that even these advanced LLMs are susceptible to self-bias amplification.

\begin{figure}[t]
  \centering
    \includegraphics[width=\linewidth]{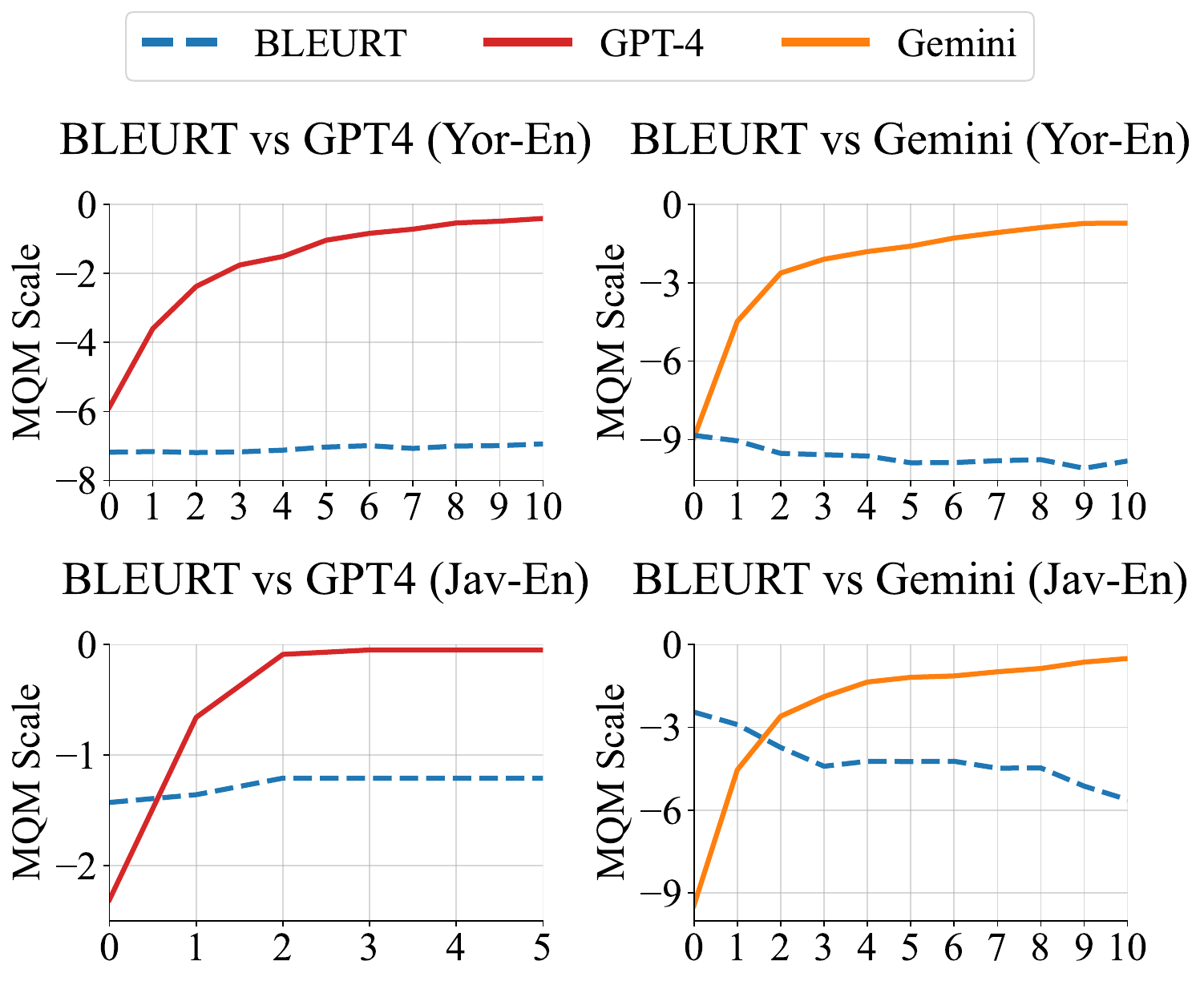}
  \caption{GPT-4 and Gemini overestimate improvements in self-refined outputs, leading to amplified bias over iterations compared to actual performance measured by BLEURT.}
\label{fig:gpt-4_gemini}
\end{figure}

In Figure \ref{fig:gpt-4_gemini}, we illustrate a comparison between GPT-4 and Gemini's quality assessments of their own outputs and performance measured by reference-based BLEURT over ten iterations. Our findings suggest that the primary reason for the amplification of bias during self-refine iteration is that actual performance does not improve through iterations. Instead, GPT-4 and Gemini mistakenly perceive performance improvements in their refined outputs. This discrepancy between the false positive performance measure and the true performance measure grows larger with each iteration. The appendix Section \ref{sec:gemini_jav_en} details Gemini's shift from right-skewed to left-skewed distribution, resulting in a decrease in distance skewness during early iterations and an increase in later ones.

\begin{figure}[t]
  \centering
    \includegraphics[width=\linewidth]{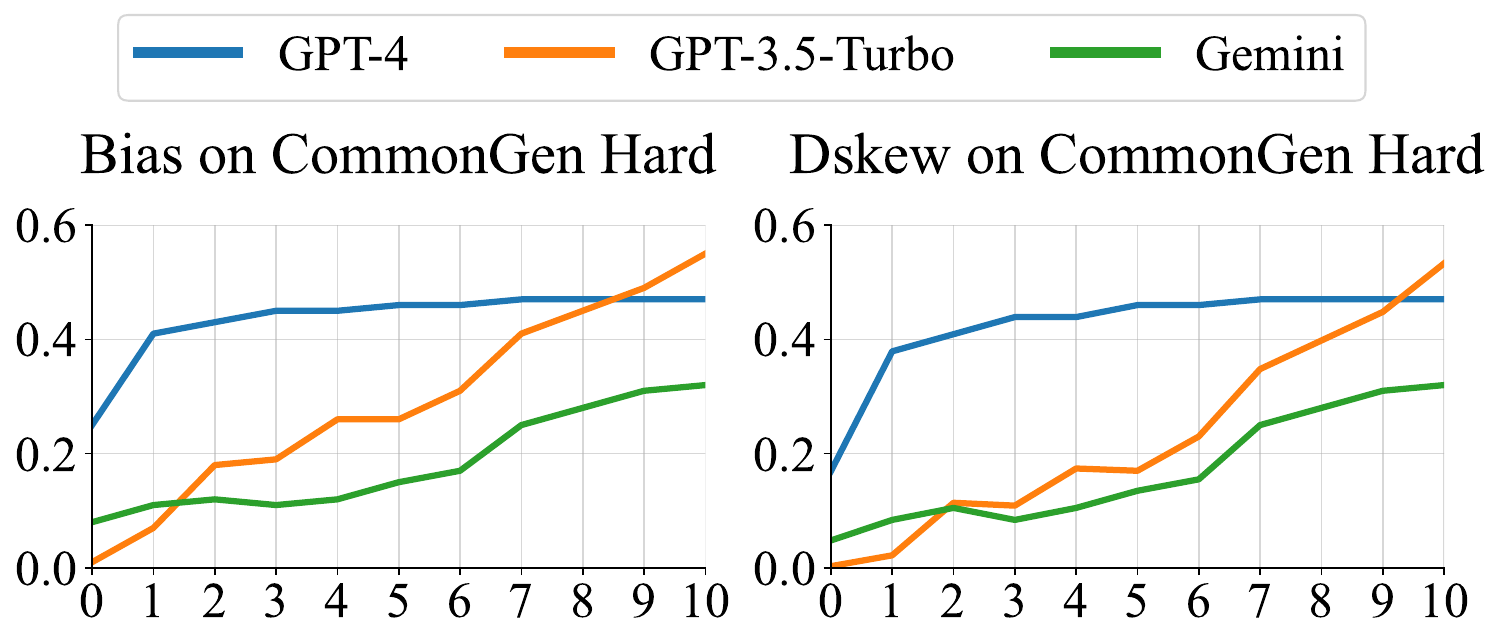}
    \includegraphics[width=\linewidth]{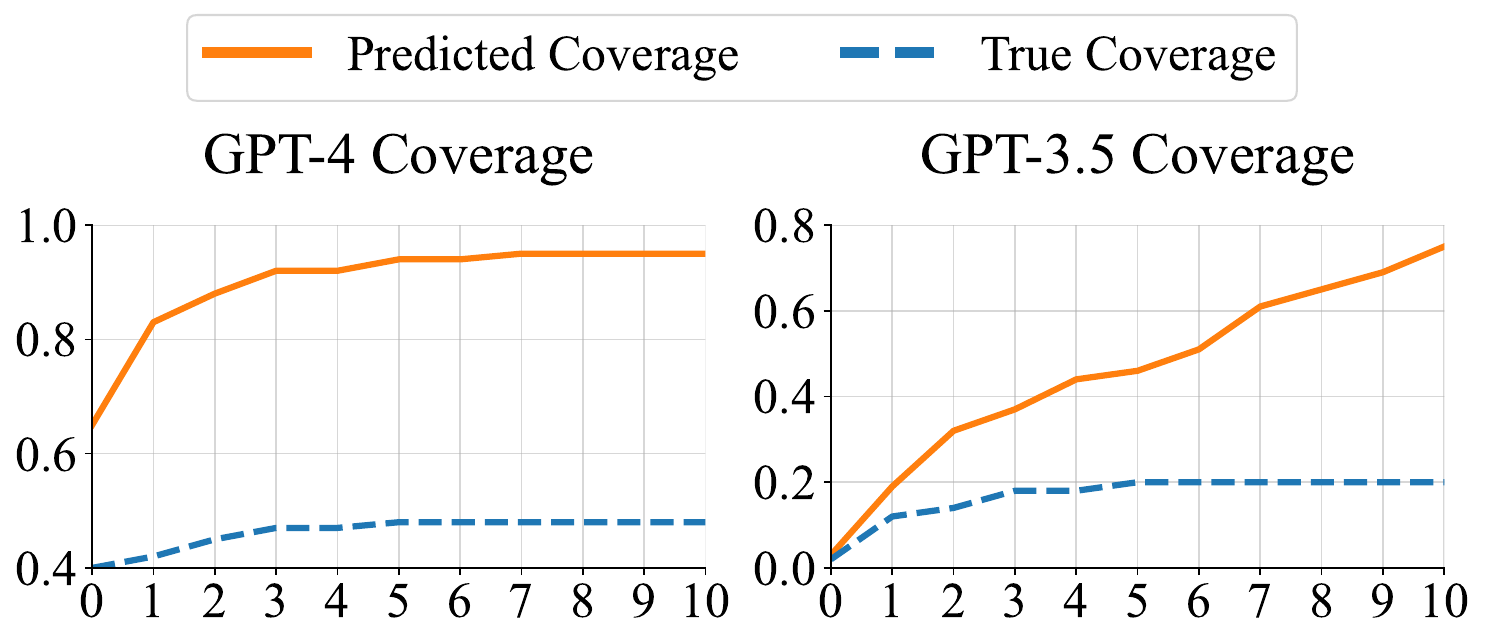}
  \caption{We evaluate the bias and distance skewness of generated texts produced by GPT-4, GPT-3.5-Turbo, and Gemini on the CommonGen dataset, across self-refinement steps. Additionally, we report the coverage of GPT-3.5-Turbo and GPT-4 compared to true concept coverage. We show that the rate of LLM's self-estimated improvements is much higher than the true coverage improvements, which leads to self-bias amplification.}
\label{fig:commongen}
\end{figure}

\paragraph{Constrained Text Generation.} Figure \ref{fig:commongen} depicts the amplification of self-bias through ten self-refine iterations in constrained text generation for GPT-3.5-Turbo, GPT-4, and Gemini. Notably, GPT-4 exhibits a higher bias estimation at earlier iterations compared to GPT-3.5-Turbo and Gemini. This can be attributed to GPT-4's higher coverage ratio at initial generation (approximately 40\%) compared to its counterparts (GPT-3.5-Turbo at around 2\%). Consequently, GPT-4 struggles to identify a few missing concepts, while GPT-3.5-Turbo and Gemini have more coverage issues and can easily identify missing input concepts.

As GPT-3.5-Turbo reaches 20\% coverage around the 5th iteration, it experiences a significant rise in bias and skewness estimation. It is worth noting that the rate of LLM's self-estimated improvements is much higher than the true coverage improvements. This phenomenon results in a saturation of performance improvements after the 5th iteration for both GPT-4 and GPT-3.5-Turbo.

\begin{figure}[ht]
  \centering
    \includegraphics[width=\linewidth]{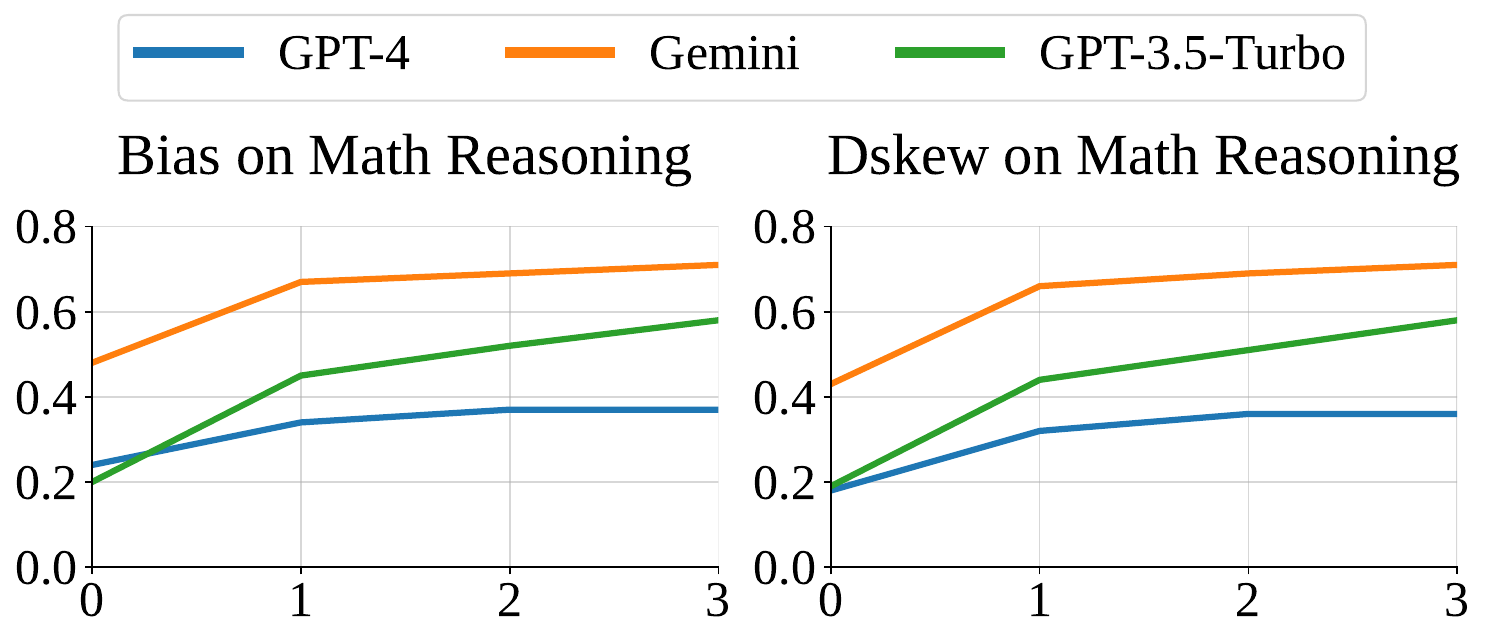}
  \caption{Bias and distance skewness in generated texts from GPT-4, GPT-3.5-Turbo, and Gemini are measured on MATH testing set throughout the self-refinement steps. Results show an increase in bias and skewness of some math solutions during iterative self-refine.}
\label{fig:math_bias_skew_self_refine}
\end{figure}

\paragraph{Mathematical Reasoning.} Figure \ref{fig:math_bias_skew_self_refine} illustrates that all large language models (LLMs) exhibit an increase in bias and skewness estimation in the iterative self-refine pipeline. This suggests that LLMs introduce self-biases towards some math solutions during self-refine.

\paragraph{Human Evaluation on Bias Estimation.} We employ one graduate student to annotate $50$ examples from the 0th and 10th iteration of GPT-4, GPT-3.5-Turbo and Gemini's outputs at Yor-En, respectively. The human rater compares candidate text against reference and labels error location, error type, and severity labels at candidate text. The scoring scheme follows MQM style \cite{freitag-etal-2021-experts}, which matches the scoring range of LLM's feedback. Our human score indicates that all three LLMs have not received measurable improvements via the self-refine pipeline (The raw human scores are included in the Appendix Table \ref{tab:human_eval_gpt-4}, \ref{tab:human_eval_gpt-3.5-turbo} and  \ref{tab:human_eval_gemini}), which is consistent with the BLEURT assessment. In Table \ref{tab:human_eval}, both increasing bias and distance skewness estimation demonstrate that all LLMs have significantly increased their self-bias with $10$ iterative refinements. In the following case study, we examine self-bias in GPT-4. Our observations reveal that GPT-4's self-feedback mechanism led to the optimization of false positives, resulting in an amplification of self-bias over three iterations. In section \ref{sec:alleviation}, we demonstrate two potential alleviation that we can use to mitigate this self-bias.

\begin{tcolorbox}
[colback=white,colframe=gray!50!black,title=Self-bias Example at GPT-4]
\small
\textbf{Yoruba text:} Ní bayii a ni àwon eku oloshu merin ti ko ni dayabetesi telele to ti ni ayabetesi,” o she afikun.\\ 
\noindent\textbf{Reference English text}: "We now have 4-month-old mice that are non-diabetic that used to be diabetic," he added.\\ 
(\textcolor{red}{Red span} indicates a major error and \textcolor{blue}{blue span} indicates a minor error annotated by GPT-4.)\\\\
\textbf{GPT-4's 1st generation} [Human: -11, GPT4: -11, Bias: 0]: "\textcolor{red}{At this point}, we have four \textcolor{red}{rats} without diabetes that have developed diabetes," \textcolor{blue}{he added}.\\\\
\textbf{GPT-4's 1st refinement} [Human: -12, GPT4: -10, Bias: 2]: "\textcolor{red}{Currently}, we have four \textcolor{red}{healthy rats} that have developed diabetes," he clarified.\\\\
\textbf{GPT-4's 2nd refinement} [Human: -11, GPT4: 0, Bias: 11]: "Presently, we have four non-diabetic rats that have developed diabetes," he elaborated.
\end{tcolorbox}

\begin{table}[ht]
\resizebox{0.49\textwidth}{!}{
 \begin{tabular}{@{}lllllll@{}}
    \toprule 
    \multicolumn{1}{c}{}
    & \multicolumn{2}{c}{GPT-4} & \multicolumn{2}{c}{GPT-3.5-Turbo} & \multicolumn{2}{c}{Gemini} \\
    \cmidrule(r){2-3}
    \cmidrule(l){4-5}
    \cmidrule(l){6-7}
    \multicolumn{1}{c}{Iterations} & Bias & Dskew & Bias & Dskew & Bias & Dskew\\
    \midrule
    \multicolumn{1}{c}{0th} & 8.06 & 0.452 & 19.6 & 0.803 & 9.62 & 0.455 \\
    \multicolumn{1}{c}{10th} & 14.6 & 0.692 & 21.9 & 0.885 & 17.6 & 0.766 \\
    \bottomrule
\end{tabular}
}
\caption{We report human evaluation on GPT-4, GPT-3.5-Turbo and Gemini's quality assessment on 0th and 10th iteration of refinement generation at Yor-En. We used Bias and Dskew estimation to demonstrate bias found by human evaluation. All LLMs have significantly increased self-bias after 10 iterations.}
 \label{tab:human_eval}
\end{table}

\paragraph{Human Evaluation on LLM's Output Quality.} We conducted human evaluation on six LLM’s self-feedback outputs at first and fifth iteration at Yoruba to English translation. For each LLM at each iteration, we annotate 100 samples. In total, we annotate 1200 samples. Specifically, human labor will check whether error annotation in the format of 'xxx' is a minor xxx error/'xxx' is a major xxx error/'xxx' is a critical xxx error (When LLM outputs an error-free annotations, it can have flexible forms, such ‘None’, ‘No error’, “Perfect translation”). 

\begin{table}[t]
\resizebox{0.50\textwidth}{!}{
 \begin{tabular}{@{}lllllll@{}}
    \toprule 
    \multicolumn{1}{c}{Iter} & Gemini & GPT3.5 & GPT4 & LLaMA2 & Mixtral & DeepS \\
    \midrule
    \multicolumn{1}{c}{1st} & 93\% & 98\% 
    & 100\% & 100\% & 100\% & 99\%\\
    \multicolumn{1}{c}{5th} & 93\% & 97\% 
    & 100\% & 100\% & 100\% & 98\%\\ 
    \bottomrule
\end{tabular}
}
\caption{We report human evaluation of format accuracy at six LLM's outputs. We observed that all LLMs have either perfect or nearly perfect format at first and fifth iteration of self-feedback at Yor-En translation. Mixtral stands for MixtralMOE and DeepS stands for DeepSeekMoE that we used in the experiment.}
 \label{tab:human_eval_format}
\end{table}

In Table \ref{tab:human_eval_format}, we include format accuracy for all LLMs. We observed that all LLMs have either perfect or nearly perfect format at first and fifth iteration of self-feedback. This is expected as we explicitly provide three in-context examples to control the output format. We found that different LLMs make different format mistakes. For example, DeepSeekMOE produces one or two garbage outputs and GPT-3.5-Turbo produces two or three free form outputs, like “The machine translation is incorrect as it provides an alternative translation that does not match the source text.” We conclude that this is due to their intrinsic instability of their instruction following capabilities. Gemini model contains surprisingly low format accuracy compared to other LLMs. This is due to the Gemini model refusing to generate any content that involves sensitive topics. There are 7 sentences in our testing set, Gemini refuses to provide responses. However, since our study focuses on self-bias amplification at iterations, this will not impact our experimental conclusions (The effects canceled out when comparing 1st and 5th iteration).

\begin{figure}[h!]
  \centering
    \includegraphics[width=\linewidth]{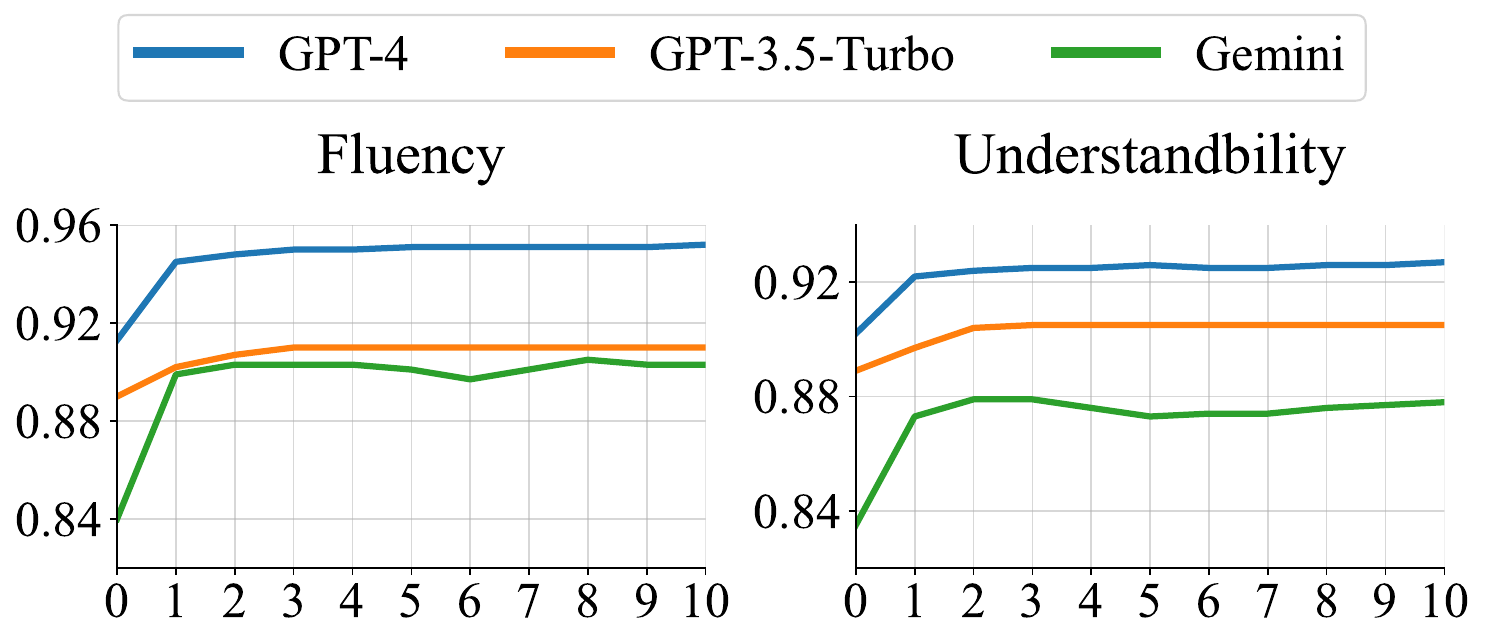}

  \caption{We measure the fluency and understandability aspects of GPT-4, GPT-3.5-Turbo, and Gemini's generated texts at Yor-En through self-refine steps. Despite no gains in quality, all LLMs have consistent performance improvements in fluency and understandability.}
\label{fig:aspect_ablation}
\end{figure}

\subsection{What improves after self-refinement?}
\paragraph{Self-refinement can improve fluency and understandability but not quality.} We demonstrate that LLM with biased feedback can impede the model's self-refine process. This raises a natural question: if an LLM does not improve its generation quality, does it improve in any other aspects throughout the iterative refine phase? To investigate this, we utilize the learned metric UniEval \cite{zhong-etal-2022-towards} to measure the LLM's improvement beyond quality metrics. UniEval, a multidimensional learned metric, estimates various evaluation dimensions, including fluency, understandability, engagement and more. We focus on two dimensions, fluency and understandability, which UniEval is not trained on task-specific data. Our results, illustrated in Figure 6, show that GPT-4, GPT-3.5-Turbo, and Gemini consistently exhibit improvements in both fluency and understandability. This suggests an alternative perspective on the self-refine pipeline, indicating that while an LLM may not strictly adhere to instruction-following in terms of quality improvements, it can still improve certain intrinsic text qualities, such as fluency and understandability.
\begin{figure}[h!]
  \centering
    \includegraphics[width=\linewidth]{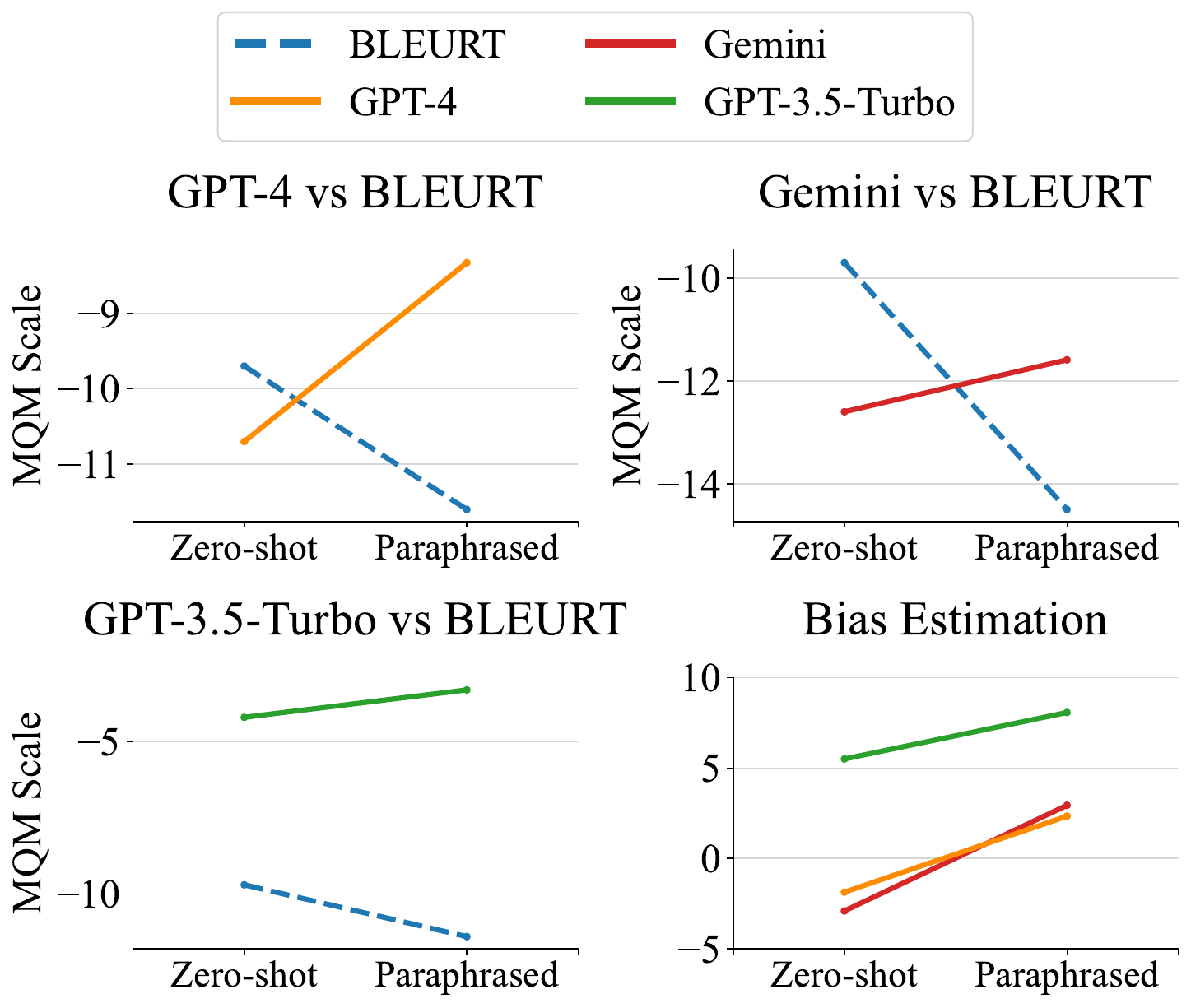}
  \caption{We used Madlad400-10b to translate 100 Yor-En translations and asked GPT-4, GPT-3.5-Turbo, and Gemini to paraphrase 100 translations. We show the BLEURT and LLM scores before and after paraphrasing. In the lower right of the figure, we show the bias estimation before and after paraphrasing. GPT-4 and Gemini have negative self-bias before paraphrasing. After paraphrasing, all LLMs increase their bias against their paraphrased outputs.}
\label{fig:paraphrase_ablation}
\end{figure}

\paragraph{LLMs favor texts that follow their style.} To explore this propensity, we conducted experiments to investigate if LLMs display a preference for outputs that align with their generation style. We asked the GPT4, GPT-3.5-Turbo, and Gemini model to paraphrase external translation outputs. In this prompt, LLMs aimed not to improve the quality of translations but rather to rewrite sentences in their corresponding styles. Using the multilingual translation system Madlad400-10b \cite{kudugunta2023madlad400}, we produced 100 Yoruba-to-English translations. Subsequently, each LLM was instructed to paraphrase the generated sentences. Our findings, shown in Figure \ref{fig:paraphrase_ablation}, reveal that GPT-4 and Gemini have negative self-bias before paraphrasing. However, after paraphrasing, all LLMs showed an increased bias against their paraphrased outputs. This is mainly attributed to a decline in quality performance post-paraphrasing, with LLMs erroneously perceiving these paraphrased outputs as indicative of improvements.

\subsection{Self-Bias is Amplified at Self-Rewarding Pipeline}
In this section, we will explore the concept of self-bias in the self-rewarding pipeline, as outlined in \cite{yuan2024selfrewarding}. The pipeline begins with an instruction fine-tuned large language model (LLM). Initially, we generate $k$ candidate responses for each input provided to the LLM. Next, the same LLM is used as a reward model to identify the best-performing candidate or to rank pairs within the collection of samples. Finally, various training objectives are applied to further train the LLM using the top-performing samples.

To illustrate the potential drawbacks of this pipeline, we carried out experiments on Yoruba to English translation task using three open-source LLMs: Deepseek-MOE, MixtralMOE, and LLaMA2-7B. For each source input, we sampled $k$ candidate responses from each model. Subsequently, we obtained self-feedback scores on these candidates employing the prompt detailed in Section \ref{sec:setup_details} and computed the corresponding self-bias. We varied $k$ across $1, 4, 8, 16$, and $32$ to examine the influence of sample size on the self-bias within the self-rewarding pipeline.

\begin{table}[t]
\resizebox{0.49\textwidth}{!}{
 \begin{tabular}{@{}lllllll@{}}
    \toprule 
    \multicolumn{1}{c}{}
    & \multicolumn{2}{c}{DeepSeekMOE} & \multicolumn{2}{c}{MixtralMOE} & \multicolumn{2}{c}{LLaMA2-7B} \\
    \cmidrule(r){2-3}
    \cmidrule(l){4-5}
    \cmidrule(l){6-7}
    \multicolumn{1}{c}{Sample Size} & Bias & Dskew & Bias & Dskew & Bias & Dskew\\
    \midrule
    \multicolumn{1}{c}{1} & 14.8 & 0.735 & 12.4 & 0.483 & 8.75 & 0.491 \\
    \multicolumn{1}{c}{4} & 16.1 & 0.795 & 10.1 & 0.490 & 14.1 & 0.580 \\
    \multicolumn{1}{c}{8} & 16.7 & 0.800 & 13.0 & 0.610 & 19.8 & 0.810 \\
    \multicolumn{1}{c}{16} & 18.0 & 0.830 & 16.9 & 0.730 & 20.7 & 0.840 \\
    \multicolumn{1}{c}{32} & 18.5 & 0.840 & 18.5 & 0.790 & 20.9 & 0.850 \\
    \bottomrule
\end{tabular}
}
\caption{We report Bias and Dskew on Deepseek-MOE, MixtralMOE and LLaMA2-7B's self-feedback with varying sample size at Yor-En. Our results indicate that both bias and distance skewness tend to increase as the sample size grows larger.}
 \label{tab:self_reward_table}
\end{table}

As shown in Table
\ref{tab:self_reward_table}, we observed that all LLMs displayed an increase in bias and distance skewness as the sample size increased. This occurs when the LLM has a biased estimation of its self-feedback, and this bias can be amplified when the sample size is increased to find the top-performing candidate according to the self-feedback. Notably, selecting samples from a larger pool, e.g. a sample size of 32, significantly increases this bias compared to selections from a smaller pool, such as a sample size of 4. When the LLM optimizes over these samples, it can further increase its self-bias and generate samples that are biased by its self-feedback.

\begin{figure}[h!]
  \centering
    \includegraphics[width=\linewidth]{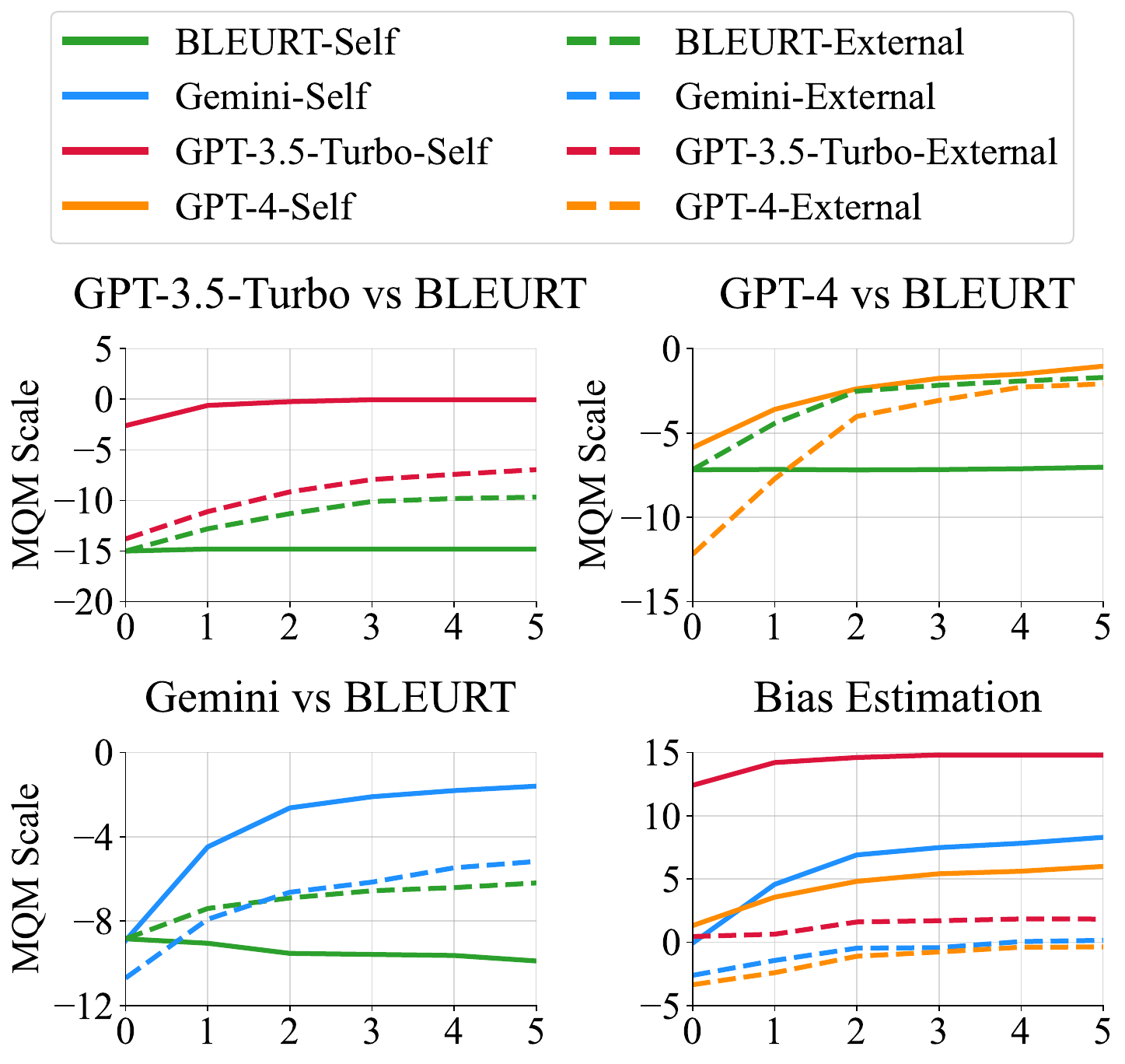}
   
  \caption{Using an external feedback model, we provide external feedback for GPT-4, GPT-3.5-Turbo, and Gemini in Yoruba-to-English translation task, across 5 refinement steps. We compare the models' true performance (measured by BLEURT) against external feedback-evaluated performance and self-feedback evaluated performance. Additionally, we plot the bias estimation for the three LLMs, considering both feedback types over 5 iterative refinement steps.}
\label{fig:explicit_ablation}
\end{figure}

\section{Alleviating Self-Bias}
\label{sec:alleviation}
\paragraph{External Feedback Reduces Self-Bias.}
We demonstrated that self-feedback from a large language model can self-amplify bias with iterative refinement. We aim to answer if external feedback with low bias estimation can improve the model's generation performance and elicit self-correction capability. We leverage a reference-based feedback model, InstructScore \cite{xu-etal-2023-instructscore}, to provide external feedback. InstructScore will take in both reference and candidate text and output fine-grained feedback, including error location, severity label, and error type. To ensure a fair comparison, we parse all outputs with the same format as self-feedback. Since InstructScore can access reference text to provide feedback, we recognize this external feedback as oracle feedback. However, models will only receive information about error location, error type, and severity labels. Therefore, refinement still relies on LLM's self-correction capability. 

In Figure \ref{fig:explicit_ablation}, we demonstrate that external feedback with accurate assessment can significantly lower the model's bias at iterative refinement (shown at the lower right of the figure. All dotted curves are below solid curves with corresponding colors). Interestingly, both Gemini and GPT-4's bias estimation is improved throughout the refinement process, as the external feedback model can over-penalize low-quality outputs. As refinement proceeds, the external feedback model converges to BLEURT quality assessment that samples achieve improved quality. Most importantly, we demonstrate that all LLMs with external feedback can elicit their self-correction ability with consistent BLEURT improvements at self-refine iterations. We include a case study example in Table \ref{tab:gpt-4_good_case}. Our finding of model improvement is consistent with prior study \cite{xu2024llmrefine} and we further demonstrate that external feedback can significantly reduce self-bias.

\begin{table}[t]
\begin{tcolorbox}
[colback=white,colframe=gray!50!black,title=External Feedback Example at GPT-4]
\small
\textbf{Yoruba text:} Ní bayii a ni àwon eku oloshu merin ti ko ni dayabetesi telele to ti ni ayabetesi,” o she afikun.\\
\textbf{Reference English text}: "We now have 4-month-old mice that are non-diabetic that used to be diabetic," he added.\\
(\textcolor{red}{Red span} indicates a major error and \textcolor{blue}{blue span} indicates a minor error annotated by GPT-4.)\\

\textbf{GPT-4's 1st generation} [Human: -11, InstructScore: -10, Bias: 1]: "At this point, we have four \textcolor{red}{rats} without diabetes that have \textcolor{red}{developed diabetes}," he added.\\

\textbf{GPT-4's 1st refinement} [Human: -2, InstructScore: -6, Bias: -4]: "\textcolor{red}{At this point}, we have four mice without diabetes that \textcolor{blue}{were diabetic}," he added.\\

\textbf{GPT-4's 2nd refinement} [Human: -1, InstructScore: -1, Bias: 0]: "We now have 4-month-old mice that are non-diabetic that were diabetic," he added.
\end{tcolorbox}
\caption{This case study demonstrates that external feedback (oracle) from InstructScore \cite{xu-etal-2023-instructscore} can remain low self-bias during iterative self-refine. By providing accurate error type, error location, and severity labels, InstructScore effectively elicits GPT-4's self-correction capability and improves its translation quality.  Despite InstructScore's oracle-like role (which it can access reference text to make error annotations), it does not provide explicit corrections, requiring GPT-4 to rely on its internal knowledge for corrections.}  
\label{tab:gpt-4_good_case}
\end{table}


\begin{figure}[h!]
  \centering
    \includegraphics[width=\linewidth]{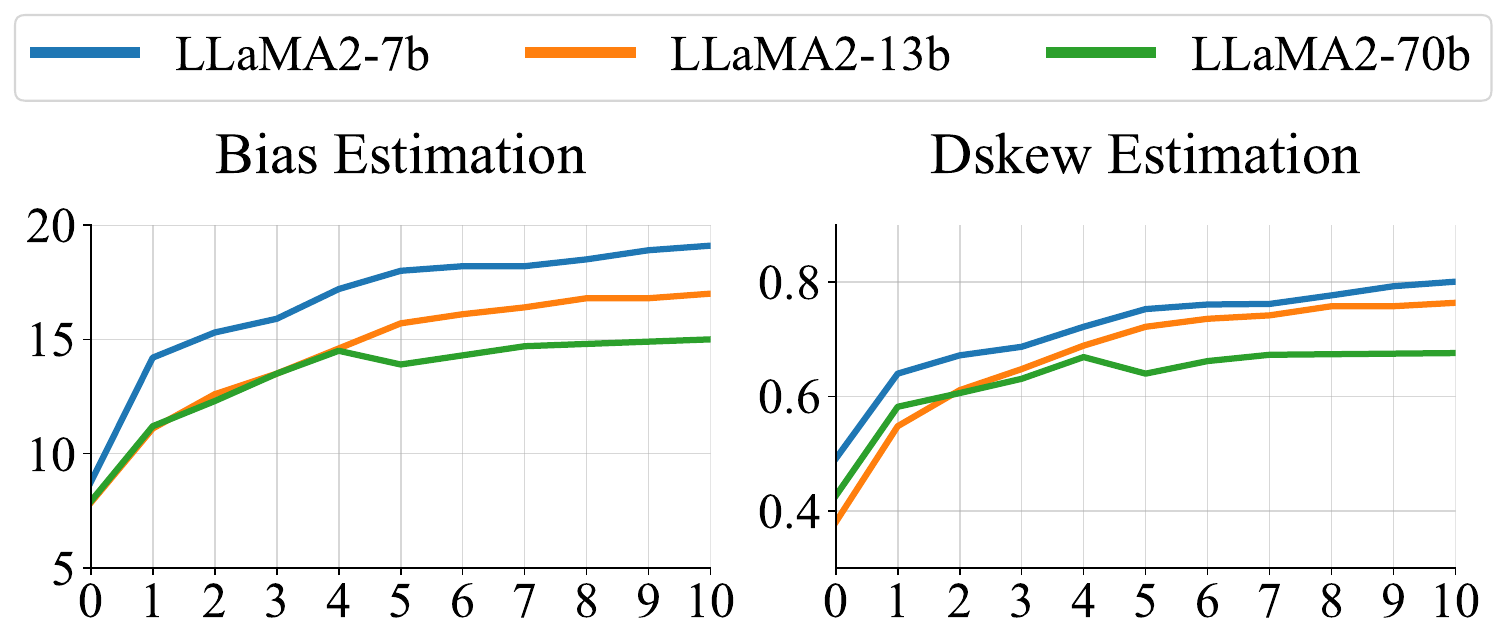}
  \caption{We show that bias and distance skewness estimation on LLaMA-2 7B, 13B, and 70B models at Yor-En translation across self-refinement steps. LLM with larger parameter size can have less self-bias.}
\label{fig:model_size_ablation}
\end{figure}

\paragraph{Larger Model Reduces Self-Bias.}
In Figure \ref{fig:model_size_ablation}, we demonstrate that LLMs with larger parameter size can have less self-bias throughout self-refinement steps. Specifically, we tested the LLaMA2 models with 7B, 13B, and 70B parameters on Yoruba-to-English (Yor-En) translation tasks. Our findings indicate that while the LLaMA2-70B model exhibits self-bias in the earlier iterations, its self-bias begins to plateau after the 5th iteration. In contrast, the 7B and 13B models continue to amplify their self-bias in later iterations. This observation aligns with prior work \cite{huang-etal-2023-large}, which posited that larger LLMs possess better self-refinement capabilities. Our study contributes to this discussion from the perspective of self-bias, proposing that larger LLMs are more resilient to self-bias. Consequently, they can assess their own outputs more accurately and possess a greater capacity for self-correction.

\section{Conclusion}
In this study, we define and quantify self-bias in LLMs with two principled estimated statistics. Our experiments across six LLM families, four languages, and three tasks reveal that self-bias is prevalent in self-refine or self-rewarding pipelines. This biased self-feedback leads to false positive objectives, hindering performance improvements during iterative refinement. Further analysis reveals that while LLM improves fluency and understanding of its generated text, they do not necessarily progress in the intended direction, such as improving quality in machine translation or expanding coverage in concept-to-word generation. Instead, LLMs tend to favor texts that adhere to their inherent styles. Finally, our research suggests that larger models are more resistant to self-bias, and incorporating external feedback significantly reduces bias, leading to performance improvements in LLMs.

\section*{Acknowledgements} This work was supported by the National Science
Foundation award \#2048122. L.L. is partly supported by a gift from Apple Inc. The views expressed are those of the author and do not reflect the official policy or position of the funding agencies. We thank Yuanjing Wei for conducting the human evaluation in our experiment.

\section*{Limitations}
In this study, we focus on quantifying the self-bias exhibited by LLMs in the self-refine pipeline. We demonstrate that self-bias will be amplified in the self-refine or self-rewarding pipeline and negatively impacts the optimization process. However, in subsequent research, it would be worthwhile to explore the measurement of bias that exists between different LLMs, as well as the bias that arises when comparing original models and their knowledge-distilled counterparts. The following questions remain open: Does LLM have more bias towards LLMs that follow the same pretraining procedure, data, or learning objectives? Does LLM have more bias to the LLMs within the same language model families? Do knowledge-distilled LLMs have more biases over the original LLMs, such as Vicuna to GPT4 or Alpaca to ChatGPT? We leave these interesting avenues for future research.



\section*{Ethical Statement}
All the benchmark data that we used during experiments is publicly available. We assure that the benchmark data does not contain risk or toxic content. The annotater was compensated fairly and did not disclose any privacy information during the annotation process. All the open sourced models can be accessed online and all the closed source models have publicly accessible APIs. The annotaters were allowed to label sensitive information if necessary. The annotater is fully aware that the data we collected from him/her will be used for research purposes. The total human annotation period took six hours and the annotator was paid above local minimum wage. We used Mistral Medium, Grammarly and ChatGPT API to polish some of our writings. 

The findings of this research have far-reaching implications for the broader linguistic and technological communities, particularly in the preservation and revitalization of endangered or low-resource languages. By identifying and mitigating self-bias in large language models (LLMs), this work paves the way for significant improvements in machine translation for languages that are underrepresented in digital platforms and datasets.

The ability to reduce bias in the self-refine pipeline of LLMs can lead to more accurate and nuanced translations, thereby enhancing the quality and accessibility of digital content in low-resource languages. This advancement is critical for preserving the cultural heritage and knowledge embodied in these languages, which are at risk of disappearing. Through improved translation capabilities, communities can more easily access global information in their native languages, fostering educational opportunities and cultural exchange. This contributes to the preservation of linguistic diversity and promotes a more inclusive digital ecosystem.
\bibliography{custom}

\newpage

\appendix

\section{Model API/Checkpoints}
\label{sec:model_ckpt}
This section provides a pointer to checkpoints that we used during experiment. All open-source models are available on the Hugging Face platform. For LLaMA2, we use "meta-llama/Llama-2-(7, 13, 70)b-chat-hf" respectively. For Mixtral MOE, we use "mistralai/Mixtral-8x7B-Instruct-v0.1". For DeepSeekMoE, we use "deepseek-ai/deepseek-moe-16b-chat". For InstructScore, we use "xu1998hz/InstructScore". For the translation model Madlad400-10b, we use "google/madlad400-10b-mt". We used GPT-3.5-Turbo and GPT-4 from OpenAI platform (https://platform.openai.com). We use gemini-pro from Google Gemini API.

\section{Quantile Mapping}
\label{sec:quantile_mapping}
While BLEURT \cite{sellam2020bleurt} correlates highly with human judgments \cite{freitag-etal-2022-results}, its scale of roughly 0 to 1 is incompatible with the MQM human annotations, which range from -25 to 0. A linear mapping is not feasible, as the BLEURT score is not calibrated to the human score, meaning a BLEURT score of 0.8 does not correspond to -5 in MQM annotations.

To address this issue, we employ quantile mapping \cite{BiasCorrectionofGCMPrecipitationbyQuantileMappingHowWellDoMethodsPreserveChangesinQuantilesandExtremes} to transform the BLEURT score into the distribution of human scores. This method involves learning a mapping function that maps the quantiles or percentiles of the predictive distribution to those of the observed distribution. In this case, our predictive distribution is derived from the BLEURT score distribution, while our observed distribution comes from the corresponding human score distribution.

We utilize the WMT22 shared metric task \cite{freitag-etal-2022-results} to obtain mapped BLEURT-human scoring pairs. In this shared metric task, each translation generated by different translation model is rated by humans using the MQM human rating scale. We also run BLEURT on the same set of translations to obtain BLEURT scores, resulting in 28125 mapped BLEURT-human scoring pairs.

We then perform the following steps: 1) Separately sort the data of the two distributions in ascending order. 2) Compute the cumulative distribution function (CDF) for each distribution. 3) Learn an interpolation function that maps the percentiles of the first distribution to the percentiles of the second distribution. 4) Apply the mapping function to the values drawn from the predictive distribution (BLEURT score distribution) to obtain the corresponding values in the observed distribution (human MQM score distribution). 

This process maps the BLEURT score distribution to the human score distribution (from -25 to 0) while preserving the relative ordering of BLEURT scores. In our experiments, we used the latest BLEURT model, BLEURT-20 checkpoint \cite{pu2021learning}, which demonstrates the highest correlation to the human judgments among its variants. 

\section{Gemini's Skewness at Translation}
\label{sec:gemini_jav_en}
Specifically, in the Java-English (Jav-En) language pair, Gemini initially assigns lower quality scores to its output compared to BLEURT assessments during early iterations, resulting in an underestimation of output performance. This phenomenon accounts for the decrease in distance skewness at the beginning, as the right-skewed distribution becomes more neutral. However, as bias accumulates in later iterations, the distribution shifts towards a left-skewed distribution, leading to an increase in distance skewness.

\begin{figure}[t]
  \centering
    \includegraphics[width=\linewidth]{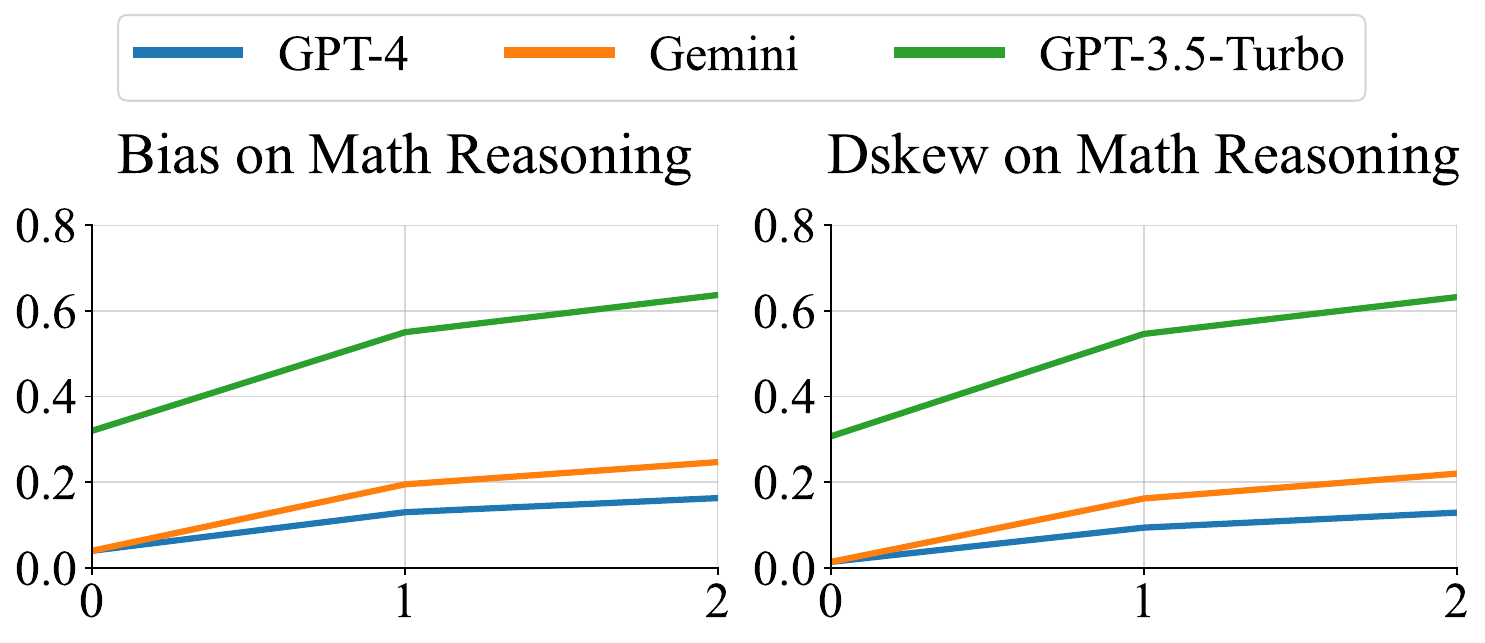}
  \caption{Bias and distance skewness in generated texts from GPT-4, GPT-3.5-Turbo, and Gemini are measured on MATH testing set throughout the self-refinement steps. Results show an increase in bias and skewness during iterative self-consistency, causing biased ensembles in reasoning paths.}
\label{fig:math_bias_skew}
\end{figure}

\section{Self-consistency results on Math reasoning}
\label{sec:self_consistency_math}
We slightly modify the self-refine pipeline by replacing the self-evaluation with self-consistency verification \cite{huang-etal-2023-large}. Namely, with the initial solution, LLM will generate an additional ten reasoning paths and a majority vote for a proposed answer. If the proposed answer is inconsistent with the prior solution, we will output a binary score of $0$, and the initial answer will be replaced by the proposed answer. Otherwise, we will output a score of $1$, and no change will be made to the initial answer. Figure \ref{fig:math_bias_skew} illustrates that all large language models (LLMs) exhibit an increase in bias and skewness estimation in the iterative self-consistency pipeline. This suggests that LLMs introduce self-biases towards certain reasoning paths during self-refine, ultimately leading to a biased ensemble across multiple reasoning paths.

\section{Additional Results}
\begin{figure*}[h!]
  \centering
    \includegraphics[width=0.99\linewidth]{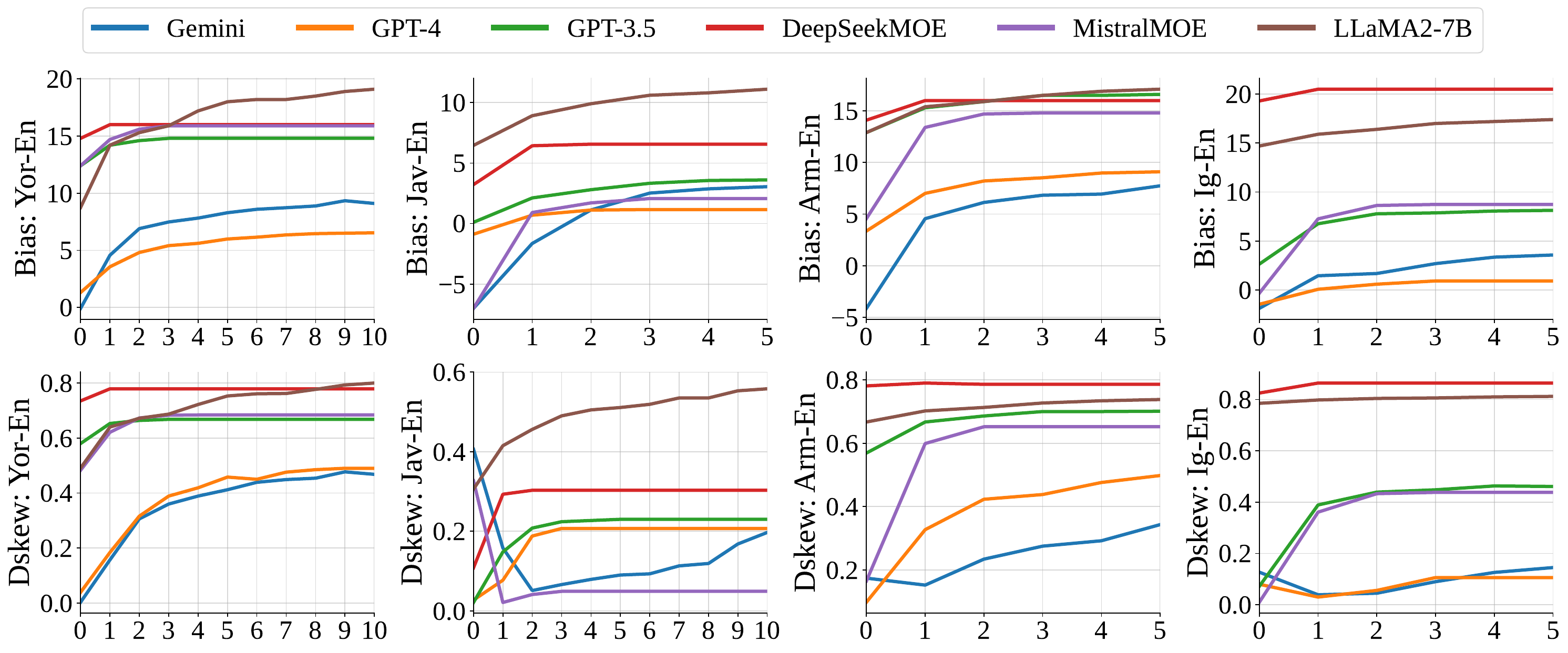}
  
  \caption{Full Bias and Dskew estimations for Yor-En, Jav-En, Arm-En, and Ig-En translations on FLores200, with the $x$-axis showing self-refine steps, reveal that all LLMs exhibit self-bias, where open-source LLMs exhibit higher levels than GPT-4 and Gemini.}
  \label{fig:bias_skewness_flores200_full}
\end{figure*}

In Table \ref{tab:human_eval_gpt-4}, we include human evaluation results and GPT-4's quality scores for the 0th and 10th iteration of refinement generation at Yorba-to-English. In Table \ref{tab:human_eval_gpt-3.5-turbo}, we include human evaluation and GPT-3.5-Turbo's quality assessment on the 0th and 10th iteration of refinement generation at Yorba-to-English. In Table \ref{tab:human_eval_gemini}, we include human evaluation and Gemini's quality assessment on the 0th and 10th iterations of refinement generation. In Figure \ref{fig:bias_skewness_flores200_full}, we include full bias and distance skewness for Yor-En, Jav-En, Arm-En and Ig-En translations on Flores200.

\label{sec:appendix}

\begin{table}[ht]
\resizebox{0.45\textwidth}{!}{
 \begin{tabular}{@{}lllll@{}}
    \toprule 
    \multicolumn{1}{c}{Human Evaluation} & Human & GPT-4 & Bias & Dskew\\ 
    \midrule
    \multicolumn{1}{c}{0th Iteration} & -15.0 & -6.92 & 8.06 & 0.452\\
    \multicolumn{1}{c}{10th Iteration} & -15.1 & -0.52 & 14.6 & 0.692\\
    \bottomrule
\end{tabular}
}
\caption{This table presents human evaluation results and GPT-4's quality scores for the 0th and 10th iteration of refinement generation performed at Yor-En. Bias and Dskew estimates are included to quantify the biases identified through human evaluation.}
 \label{tab:human_eval_gpt-4}
\end{table}

\begin{table}[ht]
\resizebox{0.45\textwidth}{!}{
 \begin{tabular}{@{}lllll@{}}
    \toprule 
    \multicolumn{1}{c}{Human Evaluation} & Human & GPT-3.5 & Bias & Dskew\\ 
    \midrule
    \multicolumn{1}{c}{0th Iteration} & -22.2 & -2.61 & 19.6 & 0.803\\
    \multicolumn{1}{c}{10th Iteration} & -21.9 & -0.03 & 21.9 & 0.885\\
    \bottomrule
\end{tabular}
}
\caption{We report human evaluation and GPT-3.5-Turbo's quality assessment on the 0th and 10th iteration of refinement generation at Yor-En. }
 \label{tab:human_eval_gpt-3.5-turbo}
\end{table}

\begin{table}[ht]
\resizebox{0.45\textwidth}{!}{
 \begin{tabular}{@{}lllll@{}}
    \toprule 
    \multicolumn{1}{c}{Human Evaluation} & Human & Gemini & Bias & Dskew\\ 
    \midrule
    \multicolumn{1}{c}{0th Iteration} & -17.3 & -8.92 & 9.62 & 0.355\\
    \multicolumn{1}{c}{10th Iteration} & -18.3 & -0.72 & 17.6 & 0.766\\
    \bottomrule
\end{tabular}
}
\caption{We report human evaluation and Gemini's quality assessment on the 0th and 10th iterations of refinement generation at Yor-En.}
 \label{tab:human_eval_gemini}
\end{table}
\begin{table*}[!ht]
\centering
\footnotesize
\begin{tabular}{p{15cm}}
\toprule
\textbf{In-context-learning prompt for LLM's initial generation at translation:}\\
\textit{Below is an instruction that describes a task. }

\#\#\# Instruction:

Translate Chinese text into English.

\begin{CJK*}{UTF8}{gbsn}
Chinese: 新华时评：把优秀返乡农民工打造成乡村振兴生力军-新华网
\end{CJK*}

\#\#\# English: Xinhua Commentary: Outstanding returning rural migrant workers can be a rural revitalization army - Xinhuanet\\\\

\textit{Below is an instruction that describes a task.}

\#\#\# Instruction:

Translate English text into German.

English: You can come back any time as our chat service window is open 24/7

\#\#\# German: Sie können jederzeit wiederkommen, da unser Chat-Service-Fenster täglich rund um die Uhr geöffnet ist\\\\

\textit{Below is an instruction that describes a task.}

\#\#\# Instruction:

Translate Yorba text into English.

Yorba: Won da Olori Skwodroni. Dilokrit Pattavee gege bi awako ofururu.

\#\#\# English: The pilot was identified as Squadron Leader Dilokrit Pattavee.\\\\

\textit{Below is an instruction that describes a task.}

\#\#\# Instruction:

Translate Yoruba text into English.

Yorba: O ko ago ilekun WiFi, O wi.\\\\

\textbf{Output for translation:}\\
Can you please turn off the WiFi, I'm done.\\

\bottomrule
\end{tabular}
\caption{Those are the translation in context learning example we used to prompt all LLMs across four language directions at Flores200. In this example, the source translation is Yourba text "O ko ago ilekun WiFi, O wi.". The English output text is from LLaMA2-7B's generation "Can you please turn off the WiFi, I'm done.".}
\label{tab:mt_init}
\end{table*}
\begin{table*}[!ht]
\centering
\footnotesize
\begin{tabular}{p{15cm}}
\toprule
\textbf{In-context-learning prompt for LLM's Self-feedback at translation:}\\
You are an annotator for the quality of machine translation. Your task is to identify errors and assess the quality of the translation.

Based on the source segment and machine translation surrounded with triple backticks, identify error types in the translation and classify them. The categories of errors are: accuracy (addition, mistranslation, omission, untranslated text), fluency (character encoding, grammar, inconsistency, punctuation, register, spelling), locale convention (currency, date, name, telephone, or time format) style (awkward), terminology (inappropriate  for context, inconsistent use), non-translation, other, or no-error.

Each error is classified as one of three categories: critical, major, and minor. Critical errors inhibit comprehension of the text. Major errors disrupt the flow, but what the text is trying to say is still understandable. Minor errors are technically errors, but do not disrupt the flow or hinder comprehension.\\\\

\begin{CJK*}{UTF8}{gbsn}
Source: ```大众点评乌鲁木齐家居商场频道为您提供高铁居然之家地址，电话，营业时间等最新商户信息，找装修公司，就上大众点评``` Translation: ```Urumqi Home Furnishing Store Channel provides you with the latest bussiness information such as the address, telephone number, bussiness hours, etc., of high-speed rail, and find a decoration company, and go to the reviews.``` Annotate errors in the translation. MQM annotations: 
\end{CJK*}\\\\

"of high-speed rail" is a critical accuracy/addition error

"go to the reviews" is a major accuracy/mistranslation error

"etc.," is a minor style/awkwards error\\\\

Source: ```I do apologise about this, we must gain permission from the account holder to discuss an order with another person, I apologise if this was done previously, however, I would not be able to discuss this with yourself without the account holders permission.``` Translation: ```Ich entschuldige mich dafür, wir müssen die Erlaubnis einholen, um eine Bestellung mit einer anderen Person zu besprechen. Ich entschuldige mich, falls dies zuvor geschehen wäre, aber ohne die Erlaubnis des Kontoinhabers wäre ich nicht in der Lage, dies mit dir involvement.``` Annotate errors in the translation. MQM annotations:\\\\

'involvement' is a major accuracy/mistranslation error

'the account holder' is a major accuracy/omission error

'wäre' is a minor fluency/grammar error

'dir' is a minor fluency/register error\\\\

Source: ```Talks have resumed in Vienna to try to revive the nuclear pact, with both sides trying to gauge the prospects of success after the latest exchanges in the stop-start negotiations.``` Translation: ```Ve Vídni se ve Vídni obnovily rozhovory o oživení jaderného paktu, přičemže obě partaje se snaží posoudit vyhlídky na úspěch po posledních výměnách v jednáních.``` Annotate errors in the translation. MQM annotations:\\\\

've Vídni' is a major accuracy/addition error

'the stop-start' is a major accuracy/omission error

'partaje' is a minor terminology/inappropriate for context error\\\\

Source: ```Talks have resumed in Vienna to try to revive the nuclear pact, with both sides trying to gauge the prospects of success after the latest exchanges in the stop-start negotiations.``` Translation: ```Ve Vídni se ve Vídni obnovily rozhovory o oživení jaderného paktu, přičemže obě partaje se snaží posoudit vyhlídky na úspěch po posledních výměnách v jednáních.``` Annotate errors in the translation. MQM annotations:\\\\

Source: ```O ko ago ilekun WiFi, O wi.``` Translation: ```He locked the WiFi door, he said.``` Annotate errors in the translation. MQM annotations:\\\\

\textbf{Output for Feedback:}\\
'He locked the WiFi door' is a critical accuracy/mistranslation error\\

\bottomrule
\end{tabular}
\caption{Those are the feedback in context learning examples (three examples) that we used to prompt all LLMs across four language directions at Flores200. In this example, the source translation is Yourba text "O ko ago ilekun WiFi, O wi.". The English candidate text is "He locked the WiFi door" from GPT-4. GPT-4's self-feedback outputs "'He locked the WiFi door' is a critical accuracy/mistranslation error".}
\label{tab:mt_feedback}
\end{table*}
\begin{table*}[!ht]
\centering
\footnotesize
\begin{tabular}{p{15cm}}
\toprule
\textbf{In-context-learning prompt for LLM's refinement at translation:}\\
\textit{Please fix all errors. You can rewrite translation if translation is bad.}\\\\

\begin{CJK*}{UTF8}{gbsn}
Source: ```大众点评乌鲁木齐家居商场频道为您提供高铁居然之家地址，电话，营业时间等最新商户信息，找装修公司，就上大众点评```
\end{CJK*}\\

Translation: ```Urumqi Home Furnishing Store Channel provides you with the latest bussiness information such as the address, telephone number, bussiness hours, etc., of high-speed rail, and find a decoration company, and go to the reviews.```\\\\

Feedback: 'of high-speed rail' is a critical accuracy/addition error\\
'go to the reviews' is a major accuracy/mistranslation error\\
'etc.,' is a minor style/awkward error\\\\

\textit{Improved Chinese-to-English translation:} \\
Dianping Urumqi Renovation and Design Channel will provide you with the address, phone number, operation time and other information of HSR Easyhome, and please come to Dianping if you are looking for a renovation company.\\\\

Source: ```I do apologise about this, we must gain permission from the account holder to discuss an order with another person, I apologise if this was done previously, however, I would not be able to discuss this with yourself without the account holders permission.```\\

Translation: ```Ich entschuldige mich dafür, wir müssen die Erlaubnis einholen, um eine Bestellung mit einer anderen Person zu besprechen. Ich entschuldige mich, falls dies zuvor geschehen wäre, aber ohne die Erlaubnis des Kontoinhabers wäre ich nicht in der Lage, dies mit dir involvement.```\\\\

Feedback:\\
'involvement' is a major accuracy/mistranslation error\\

'the account holder' is a major accuracy/omission error\\

'wäre' is a minor fluency/grammar error\\

'dir' is a minor fluency/register error\\\\

\textit{Improved English-to-German translation:}\\
Ich bitte um Entschuldigung, aber wir benötigen das Einverständnis des Kontoinhabers, um eine Bestellung mit einer anderen Person zu besprechen, falls es schon eingeholt wurde, entschuldige ich mich, aber ich kann dies ohne das Einverständnis des Kontoinhabers nicht mit Ihnen besprechen.\\

Source: ```Talks have resumed in Vienna to try to revive the nuclear pact, with both sides trying to gauge the prospects of success after the latest exchanges in the stop-start negotiations.```\\
Translation: ```Ve Vídni se ve Vídni obnovily rozhovory o oživení jaderného paktu, přičemže obě partaje se snaží posoudit vyhlídky na úspěch po posledních výměnách v jednáních.```\\\\

Feedback:\\
've Vídni' is a major accuracy/addition error\\
'the stop-start' is a major accuracy/omission error\\
'partaje' is a minor terminology/inappropriate for context error\\\\

\textit{Improved English-to-Cezch translation:}\\
Ve Vídni byly obnoveny rozhovory o oživení jaderného paktu a obě strany se snaží odhadnout, jaké jsou vyhlídky na úspěch po posledních výměnách názorů v rámci přerušených jednání.\\\\

Source:  “‘O ko ago ilekun WiFi, O wi.“‘\\
Translation: ‘He locked the WiFi door, he said.“‘\\\\

Feedback:\\
’He locked the WiFi door’ is a critical accuracy/mistranslation error\\\\

\textit{Improved English-to-Yoruba translation:}\\
\textbf{Output for refinement:}\\
"He secured the WiFi connection, he said."\\

\bottomrule
\end{tabular}
\caption{Those are the refinement in context learning examples (three examples) that we used to prompt all LLMs across four language directions at Flores200. In this example, the source translation is Yourba text "O ko ago ilekun WiFi, O wi.". The English text is GPT-4's output "He locked the WiFi door.". We gathered GPT-4's self-feedbak: "’He locked the WiFi door’ is a critical accuracy/mistranslation error". GPT-4 outputs a refined output: "He secured the WiFi connection, he said.", according to the feedback and prior step generation.}
\label{tab:mt_refinement}
\end{table*}
\begin{table*}[!ht]
\centering
\footnotesize
\begin{tabular}{p{15cm}}
\toprule
\textbf{Prompt for LLM's initial generation for reasoning:}\\

You are a competitive math problem solver. Please generate a step-by-step solution. Your final answer must be enclosed in LaTeX's boxed tag.\\\\

How many vertical asymptotes does the graph of $y=\frac{2}{x^2+x-6}$ have?\\\\

\textbf{Output for math answer:}\\
A vertical asymptote is a vertical line that the graph of a function approaches but never crosses. A function has a vertical asymptote at $x = k$ if it has infinite limits as $x$ approaches $k$ from the left or right.For the rational function $y = \frac{2}{x^2 + x - 6}$, we know that it could have vertical asymptotes wherever the denominator is zero, because division by zero is undefined and yields infinity.Therefore, to find the vertical asymptotes, we solve the equation $x^2 + x - 6 = 0$. We apply factorization and get $(x - 2)(x + 3) = 0$. This gives us $x = 2$ and $x = -3$.In any case, we need to check if any of these could be holes instead of vertical asymptotes. This would be the case if $x = 2$ or $x = -3$ also made the numerator zero. However, the numerator is a constant and won't be zero. Therefore this function has two vertical asymptotes at $x=2$ and $x=-3$. The answer is $\boxed{2}$.\\

\bottomrule
\end{tabular}
\caption{This is the prompt we used to generate mathematical reasoning. The output in this example comes from GPT-4. We first used this prompt to generate one reasoning path as initial generation. Then, we sample $10$ different reasoning paths and use majority vote to find the most confident answer. To generate self-feedback, we compare majority voted answer and initial answer. If they are matched, feedback indicates answer is correct and no change should be made. If they are not matched, the feedback indicates answer is wrong and inital answer will be replaced by the majority voted answer. We alternate above two steps and iteratively refine the reasoning paths. }
\label{tab:math_init}
\end{table*}
\begin{table*}[!ht]
\centering
\footnotesize
\begin{tabular}{p{15cm}}
\toprule
\textbf{Concept-to-text prompt for LLM's initial generation:}\\
You are generating text based on specified words. Do not provide any explanations or text apart from the text output.\\\\

Please generate a sentence that contains the exact string matches for the following concepts:\\\\

['vest', 'snow', 'remove', 'silhouette', 'dunk', 'step', 'item', 'toddler', 'plaster', 'wall', 'piece', 'tool', 'circle', 'pull', 'wrap', 'headphone', 'deck', 'suit', 'press', 'mother', 'pit', 'cupcake', 'priest', 'rink', 'counter', 'rest', 'piano', 'wine', 'serve', 'winter']\\\\

\textbf{Output for Concept-to-Text generation:}\\\\
In the winter, the toddler wearing a vest, pressed a piece of wet plaster against the wall in a circle, while his mother baking a cupcake was removing the snow from the deck with a tool; switching steps to the rhythm of a piano piece echoing from the headphones wrapped around her neck, she caught a glimpse of the priest serving wine, his silhouette dunking a cupcake into a pit at the rink counter, under the restful gaze of a suit-clad man.\\

\bottomrule
\end{tabular}
\caption{This is the concept-to-text prompt we used to test all LLMs at Commongen Hard. This example contains the output from GPT-4.}
\label{tab:commongen_init}
\end{table*}
\begin{table*}[!ht]
\centering
\footnotesize
\begin{tabular}{p{15cm}}
\toprule
\textbf{Concept-to-text in context learning prompt for LLM's self-feedback:}\\
We want to create a sentence that contains all the specified concepts. Please provide feedback on the following sentences. The feedback should list all missing concepts. If all concepts are covered, output 'all covered'\\\\

Concepts: ['dog', 'frisbee', 'catch', 'throw']\\
Generated Sentence: A dog leaps to catch a thrown frisbee.\\
\textit{Feedback:} all covered\\\\

Concepts: ['dog', 'frisbee', 'catch', 'throw']\\
Generated Sentence: Two dogs are throwing frisbees at each other .\\
\textit{Feedback:} ['catch']\\\\

Concepts: ['vest', 'snow', 'remove', 'silhouette', 'dunk', 'step', 'item', 'toddler', 'plaster', 'wall', 'piece', 'tool', 'circle', 'pull', 'wrap', 'headphone', 'deck', 'suit', 'press', 'mother', 'pit', 'cupcake', 'priest', 'rink', 'counter', 'rest', 'piano', 'wine', 'serve', 'winter']\\
Generated Sentence: In the winter, the toddler wearing a vest, pressed a piece of wet plaster against the wall in a circle, while his mother
baking a cupcake was removing the snow from the deck with a tool; switching steps to the rhythm of a piano piece
echoing from the headphones wrapped around her neck, she caught a glimpse of the priest serving wine, his silhouette
dunking a cupcake into a pit at the rink counter, under the restful gaze of a suit-clad man.\\
\textit{Feedback:}\\\\
\textbf{Output for Concept-to-Text feedback:}\\
all covered\\

\bottomrule
\end{tabular}
\caption{This is the in-context learing (ICL) concept-to-text prompt (two ICL examples) we used to generate LLM's self-feedback at Commongen Hard. This example contains the self-feedback from GPT-4.}
\label{tab:commongen_feedback}
\end{table*}
\begin{table*}[!ht]
\centering
\footnotesize
\begin{tabular}{p{15cm}}
\toprule
\textbf{Concept-to-text prompt for LLM's refinement:}\\
Concepts: ['fruit', 'motorcycle', 'perform', 'jacket', 'vehicle', 'place', 'mat', 'walk', 'world', 'area', 'kiss', 'mother', 'pass', 'report', 'club', 'axis', 'tricep', 'patient', 'listen', 'owner', 'uniform', 'floor', 'hamburger', 'use', 'wine', 'cross', 'bull', 'sell', 'lawn', 'friend']\\
Generated sentence: In a world where a fruit can perform like a motorcycle, I found my mother wearing a leather jacket, passing by her vehicle to walk towards the place where she runs a club. Not far from that area, my friend, a patient person, was ready to sell mats, used for gym workouts, especially for triceps. Being the owner of the shop, he always wears a uniform. While I was listening to this report, sitting cross-legged on the floor and munching my hamburger, I missed seeing a bull wandering on the lawn just outside. Just then, I received an unexpected kiss from my wife who brought me a glass of wine from our cellar which also acts as an axis for all the happenings of my life.\\
Missing Concepts: ['use', 'lawn']\\
Please revise generated sentence which covers all missing concepts and all given concepts, with the exact string matches. New sentence:\\\\

\textbf{Output for Concept-to-Text refinement:}\\
In a world where a fruit can perform like a motorcycle, I found my mother wearing a leather jacket, passing by her vehicle to walk towards the place where she runs a club. Not far from that area, my friend, a patient person, was ready to sell mats, used for gym workouts, especially for triceps. Being the owner of the shop, he always wears a uniform. While I was listening to this report, sitting cross-legged on the floor and munching my hamburger, I missed seeing a bull wandering on the lawn just outside. Just then, I received an unexpected kiss from my wife who brought me a glass of wine from our cellar which also acts as an axis for all the happenings of my life.\\

\bottomrule
\end{tabular}
\caption{This is the concept-to-text refinement prompt we used to generate LLM's refinement at Commongen Hard. This example contains the refinement from GPT-4.}
\label{tab:commongen_refine}
\end{table*}
\end{document}